\documentclass{m3ailpreprint}

\usepackage{amsmath,amsfonts,bm}

\def\eqref#1{equation~\ref{#1}}

\def\1{\bm{1}}

\DeclareMathAlphabet{\mathsfit}{\encodingdefault}{\sfdefault}{m}{sl}
\SetMathAlphabet{\mathsfit}{bold}{\encodingdefault}{\sfdefault}{bx}{n}

\usepackage{amsmath,amssymb,amsfonts}
\usepackage[mono]{inconsolata}
\usepackage{booktabs}
\usepackage{colortbl}
\usepackage{enumitem}
\usepackage{float}
\usepackage{graphicx}
\usepackage{capt-of}
\usepackage{wrapfig}
\usepackage{array}
\usepackage{tabularx}
\usepackage{microtype}
\usepackage{multirow}
\usepackage{xcolor}
\usepackage{url}

\colorlet{paperlink}{mthreeblue}
\definecolor{tableOurs}{HTML}{E1EFF7}
\definecolor{qualityBlue}{HTML}{E7F0F8}
\definecolor{qualityInk}{HTML}{264D63}
\hypersetup{
  colorlinks=true,
  citecolor=paperlink,
  linkcolor=purple,
  urlcolor=paperlink
}

\newcommand{\tablecell}[2][c]{%
  {\setlength{\tabcolsep}{0em}%
   \begin{tabular}[#1]{c}#2\end{tabular}}%
}

\newcommand{\yes}{\ensuremath{\checkmark}}

\renewcommand{\paragraph}[1]{\textbf{#1}}

\title{SparseEngine: Sparse-First Inference Engine} 

\author[*]{Jitai Hao}
\author[*]{Quansheng Gu}
\author[\dagger]{Qiang Huang}
\author[\dagger]{Jun Yu}

\affiliation{Harbin Institute of Technology (Shenzhen)}

\contribution{$^*$Equal contribution.}
\contribution{$^\dagger$Corresponding authors.}

\checkdata[Email]{\email{jitaihao@outlook.com}, \email{gqs060905@outlook.com}, \email{huangqiang@hit.edu.cn}, \email{yujun@hit.edu.cn}}

\abstract{Long-context LLM agents accumulate tool outputs and reasoning traces across turns, increasing KV-cache memory, limiting concurrency, and slowing decoding.
Sparse inference reduces these costs, but engines tied to specific layouts or workflows struggle to support heterogeneous cache representations, metadata, and update rules.
The central challenge is to accommodate this diversity while preserving shared scheduling, capacity management, and cache reuse.
We introduce \textbf{SparseEngine}, a ground-up, sparse-first inference engine whose shared lifecycle contract combines fine-grained hooks and method-owned cache management to preserve control over KV representations, computation, and state transitions within shared serving infrastructure.
It supports 15 cache methods spanning dynamic sparsity, KV eviction, compression, and quantization across diverse architectures, including hybrid attention models.
The contract also enables cross-request management: \textbf{Chain Cache} resumes eviction-based methods across turns from compacted KV and method metadata without prefix reconstruction; \textbf{Prefix Cache Pruning} reclaims KV from application-selected regions while preserving logical-prefix matching.
Evaluations show that SparseEngine closely preserves reference-method quality.
On 128K-token inputs, it delivers 9.8--12.1$\times$ vanilla vLLM's aggregate decode throughput with KV eviction at each configuration's largest successfully tested batch and over 2.5$\times$ with dynamic sparsity at matched concurrency.
Multi-turn agent replay achieves up to 2.24$\times$ end-to-end speedup over full attention.
The code is available at \url{https://github.com/CURRENTF/SparseEngine}.
}

\begin{document}

\maketitle

\section{Introduction}
\label{sec:introduction}

\begin{wrapfigure}{R}{0.6\textwidth}
  \vspace{-2.0em}
  \centering
  \includegraphics[width=0.99\linewidth]{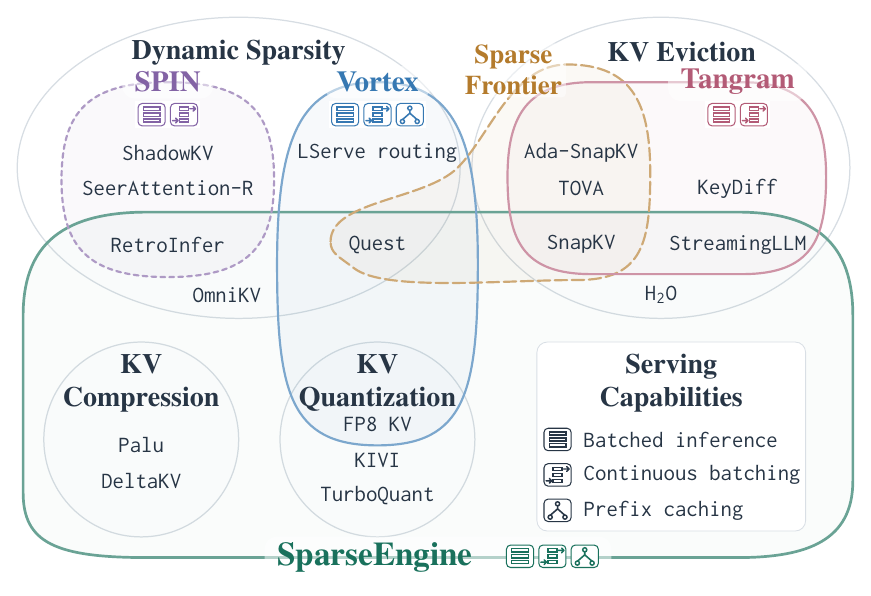}
  \vspace{-1.0em}
  \caption{\textbf{Representative method coverage and serving capabilities of sparse inference engines.}
  Gray outlines mark method families; colored outlines show each engine's coverage, with overlaps marking shared methods. Icons beside engine names show system-level support for batched inference, continuous batching, and prefix caching.}
  \label{fig:system-comparison}
  \vspace{-0.5em}
\end{wrapfigure}

As LLMs evolve from single-turn generators into autonomous, multi-turn agents, inference becomes fundamentally stateful~\citep{Yao2022ReActSR, Shinn2023ReflexionLA}. 
Agent workflows such as coding, research, and reasoning repeatedly invoke the model, call tools, and append observations and reasoning traces to an expanding context. 
This growth creates two coupled GPU bottlenecks: the KV cache consumes increasing memory and limits serving concurrency, while attention over the accumulated history increases decoding latency.

Sparse inference methods address these bottlenecks through several complementary strategies.
Dynamic sparsity accesses only query-relevant context; KV eviction discards selected entries; KV compression replaces full-dimensional KV with compact representations; and KV quantization reduces storage precision~\citep{Tang2024QuestQS, Li2024SnapKVLK, Chang2024PaluCK, Liu2024KIVIAT, Hao2025OmniKVDC}.
However, these methods differ not only in which context they retain or access, but also in their physical representations, persistent metadata, update schedules, and execution paths.
For example, Quest performs query-dependent page selection, whereas H$_2$O maintains cumulative attention scores and physically evicts KV throughout decoding~\citep{Tang2024QuestQS, Zhang2023H2OHO}.
A serving engine must accommodate these differences while still providing shared capabilities such as request scheduling, continuous batching, batched execution, capacity management, and prefix reuse.

This tension is an abstraction-design problem.
Programming languages illustrate the underlying trade-off: MATLAB makes matrix computation concise~\citep{Moler2020HistoryMATLAB}, SQL provides a declarative interface for relational queries~\citep{Astrahan1975SEQUEL}, and Rust enforces memory safety through ownership and borrowing~\citep{Jung2018RustBelt}.
Each abstraction makes some computations natural by imposing assumptions on data and state, but those assumptions can constrain other computational patterns.
Likewise, an inference-engine interface that assumes a particular KV layout or update workflow can simplify compatible sparse methods while forcing fundamentally different methods to reformulate their state or computation.

Existing sparse inference engines embody different choices along this design space.
Vortex exposes programmable page-centric operations~\citep{Chen2026VortexEA}; SPIN maps sparse methods to partitions managed through a hierarchical GPU-CPU pipeline~\citep{Zhao2026UnifyingSA_SPIN}; and Tangram targets non-uniform head-wise KV retention~\citep{Kim2026TangramUN}.
These abstractions efficiently support their intended access or retention patterns, but their assumptions limit methods requiring different storage representations, persistent state, or update sequences.
Figure~\ref{fig:system-comparison} compares representative method coverage across engines alongside their serving capabilities.
It motivates the following question:

\par
\vspace{0.3em}
\begingroup
\setlength{\fboxsep}{0em}%
\noindent\colorbox{paperlink!4}{%
  {\color{paperlink!75}\vrule width 2pt}%
  \hspace{0.8em}%
  \begin{minipage}{\dimexpr\linewidth-4pt-1.6em\relax}
    \vspace{0.7em}
    Where should the abstraction boundary lie between shared serving infrastructure and method-specific control over KV state and computation?
    \par\vspace{0.7em}
  \end{minipage}%
  \hspace{0.8em}%
}\par
\vspace{0.3em}
\endgroup

We introduce \textbf{SparseEngine}, a ground-up, sparse-first inference engine that places this boundary at a \textbf{shared lifecycle contract}.
Fine-grained hooks let each sparse method define its computation and state transitions across prefill, attention, layer, and decoding boundaries.
Method-owned cache managers control KV representations, layouts, metadata, and updates, while common interfaces connect this state to attention execution, capacity management, and scheduling.
This design integrates \textbf{15 methods} spanning dynamic sparsity, KV eviction, KV compression, and KV quantization without requiring method-specific model implementations.
It also supports diverse model architectures, including GQA, MLA, MoE, and hybrid models that combine linear and grouped-query attention.
SparseEngine further extends sparse state management across requests.
\textbf{Chain Cache} preserves compacted KV and method-specific metadata, allowing eviction-based methods to resume across agent turns without reconstructing the complete prefix.
\textbf{Prefix Cache Pruning} lets applications reclaim physical KV from selected history regions while preserving the logical prefix used for matching and reuse.

Our evaluation shows that this generality does not sacrifice efficiency or method fidelity.
Across 20 LongBench V1 and V2 comparisons, SparseEngine differs from the reference implementations by only +0.17 points on average.
On 128K-token inputs, SnapKV achieves over \textbf{10$\times$} the aggregate decode throughput of vanilla vLLM when each configuration runs at its largest successfully tested batch size, while Quest and OmniKV reach over \textbf{2.5$\times$} vLLM throughput at matched concurrency.
On multi-turn agent workloads, Chain Cache preserves quality under appropriate sparsity budgets and achieves up to \textbf{2.24$\times$} end-to-end replay speedup, while application-aware Prefix Cache Pruning reclaims tool-output KV and recovers full-attention performance by retaining recent evidence.

Our contributions are summarized as follows:
\begin{itemize}[left=1pt,] 
  \item \textbf{A lifecycle abstraction for heterogeneous sparse inference.}
  Fine-grained hooks and method-owned cache state integrate 15 methods across four families and diverse model architectures while retaining shared serving capabilities.
  
  \item \textbf{Cross-request sparse-state management.}
  Chain Cache reuses compacted method state across agent turns, while Prefix Cache Pruning reclaims selected physical KV without forfeiting logical-prefix matching.
  
  \item \textbf{Efficient serving with faithful method quality.}
  SparseEngine translates sparse computation and KV-memory savings into higher decode throughput and faster multi-turn agent replay while closely preserving reference-method quality.
\end{itemize}

\section{Related Work}
\label{sec:related-work}


\paragraph{Sparse Attention.}
KV-cache optimizations follow four complementary strategies.
\textbf{Dynamic sparse attention}, including Quest, InfLLM, and OmniKV~\citep{Tang2024QuestQS, Xiao2024InfLLMTL, Hao2025OmniKVDC}, select query-dependent context while typically retaining unselected history for future queries. 
NSA~\citep{Yuan2025NativeSA} learns native sparse-attention patterns, whereas IndexCache~\citep{Bai2026IndexCacheAS} reuses indices across layers.
\textbf{KV eviction}, including H$_2$O, SnapKV, and PyramidKV~\citep{Zhang2023H2OHO, Li2024SnapKVLK, Cai2024PyramidKVDK}, discard entries to reduce memory and subsequent attention work.
\textbf{KV compression}, including Palu, LoRC, and DeltaKV~\citep{Chang2024PaluCK, Zhang2024LoRCLC, Hao2026DeltaKVRK}, replaces full-dimensional KV tensors with compact representations; \textbf{KV quantization}, including KIVI and TurboQuant~\citep{Liu2024KIVIAT, Zandieh2025TurboQuantOV}, reduces precision.
Both alter stored payloads and may require specialized metadata or reconstruction.

These methods optimize different aspects of inference: which KV entries attention reads, which entries the cache retains, and how retained entries are stored. 
\textbf{SparseEngine}'s shared lifecycle contract integrates them with common scheduling and batching, enables method control over KV state and attention execution, and compatible combinations of selection, eviction, and quantization.

\paragraph{Inference Engine.}
Inference engines coordinate model execution, scheduling, and KV-cache management.
Continuous batching updates active requests between decoding steps, while chunked prefill divides prompts into smaller scheduling units~\citep{Yu2022OrcaAD, Agrawal2024TamingTT}.
Paged caches map logical token positions to physical KV blocks~\citep{Kwon2023EfficientMM}; prefix caching reuses computed state across requests with shared prefixes~\citep{Zheng2023SGLangEE}.

Sparse inference systems additionally coordinate sparse execution, data placement, and memory reclamation.
\textbf{SparseFrontier}~\citep{Nawrot2025TheSF} evaluates accuracy-efficiency trade-offs.
\textbf{SPIN}~\citep{Zhao2026UnifyingSA_SPIN} unifies selection granularities through partitions backed by hierarchical GPU-CPU KV storage; \textbf{Vortex}~\citep{Chen2026VortexEA} provides programmable page-centric routing; and \textbf{Tangram}~\citep{Kim2026TangramUN} supports non-uniform head-wise retention through budget reservation and Ragged Paging.
\textbf{HiSparse}~\citep{Xie2026HiSparseSS} retains full KV history on the host and fetches selected entries into a bounded GPU cache, whereas \textbf{vToken}~\citep{Gao2026vTokenTV} reclaims token-level storage through logical-to-physical indirection and repacking.

These systems target specific access, placement, or retention patterns.
\textbf{SparseEngine} instead exposes scheduling and capacity requirements through shared lifecycle interfaces while leaving physical representations and execution to each method.
This supports heterogeneous methods with continuous batching and method-compatible prefix caching (Figure~\ref{fig:system-comparison}).
Across requests, Chain Cache preserves compacted KV and metadata across turns, while Prefix Cache Pruning reclaims selected history without forfeiting logical-prefix reuse.
Appendix~\ref{app:background} provides detailed background and notation.

\section{SparseEngine}
\label{sec:method}

\paragraph{Lifecycle Contract.}
Existing sparse inference engines integrate methods through predefined workflows.
SPIN~\citep{Zhao2026UnifyingSA_SPIN} follows ``$\text{Index} \!\rightarrow\! \text{Offload} \!\rightarrow\! \text{Select} \!\rightarrow\! \text{Retrieve} \!\rightarrow\! \text{Attention}$,'' whereas Vortex~\citep{Chen2026VortexEA} follows ``$\text{Cache Update \& Summary Calculation} \!\rightarrow\! \text{Page Selection} \!\rightarrow\! \text{Attention}$.''
SparseEngine instead anchors its contract to a \textbf{key invariant}: the native module execution order of a given model.
Fine-grained hooks along this order let sparse methods register callbacks and compose custom workflows without modifying model implementations.

Section~\ref{sec:method:architectural-foundations} develops three components that realize this contract:
(1) \texttt{SparseController} exposes lifecycle hooks;
(2) a method-customized \texttt{CacheManager} owns KV storage and persistent metadata; and
(3) \texttt{AttentionView} connects method-owned state to attention execution.
Section~\ref{sec:method:cache-state-management} then extends this contract across requests: Chain Cache reuses retained KV and method state across agent turns, while Prefix Cache Pruning reclaims physical KV from selected history regions without losing logical-prefix matching or reuse.

\begin{figure}[t]
  \centering
  \includegraphics[width=0.99\textwidth]{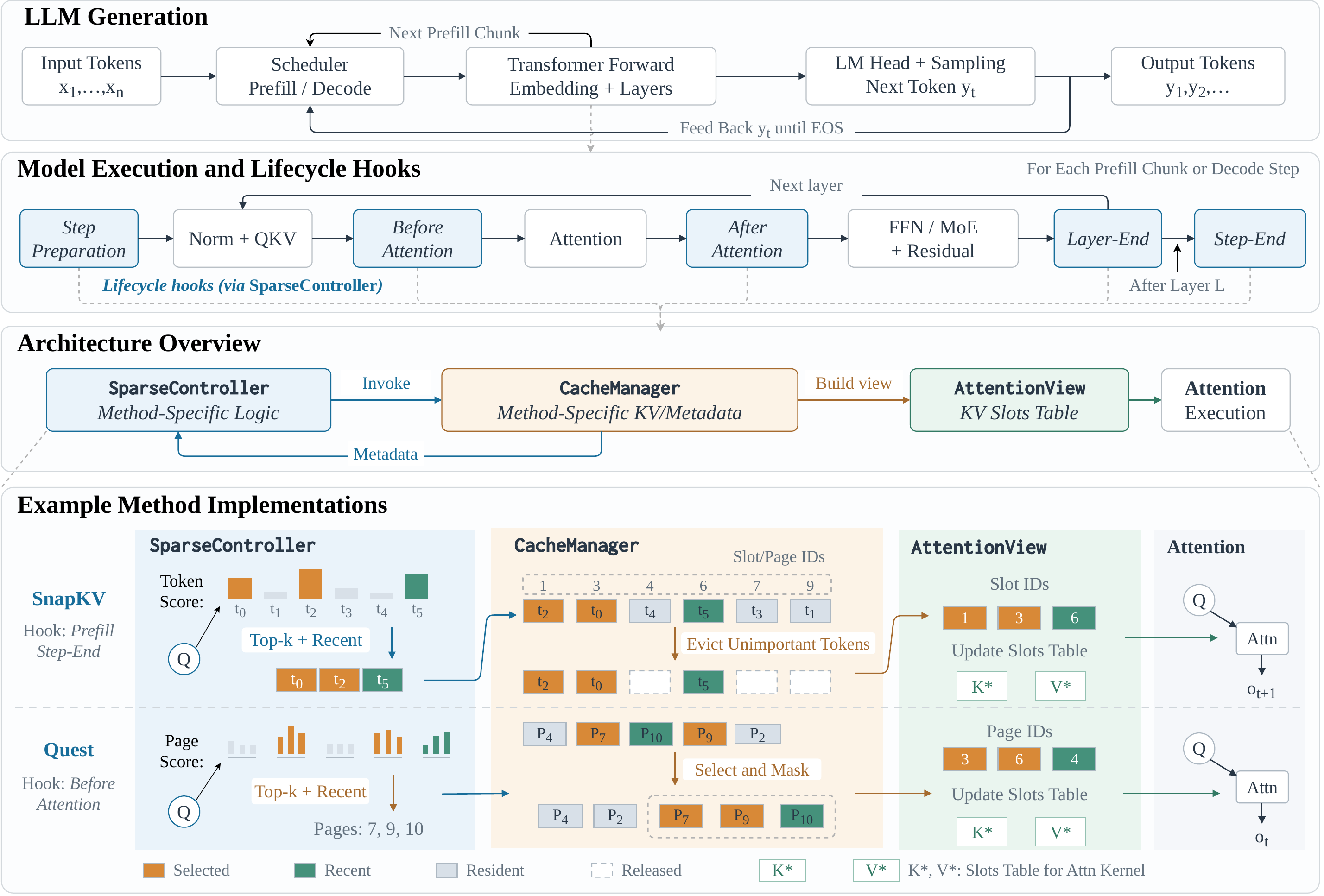}
  \vspace{-0.5em}
  \caption{\textbf{Overview of SparseEngine.}
  From top to bottom, the figure shows the LLM generation loop; the Transformer execution path and lifecycle hooks for prefill and decoding; the interactions among \texttt{SparseController}, the method-specific \texttt{CacheManager}, \texttt{AttentionView}, and attention execution; and example implementations of SnapKV token eviction and Quest query-dependent page selection.}
  \label{fig:execution-overview}
  \vspace{-0.5em}
\end{figure}

\subsection{Architectural Foundations}
\label{sec:method:architectural-foundations}

\paragraph{Fine-Grained Lifecycle Hooks in \texttt{SparseController}.}
Figure~\ref{fig:execution-overview} shows how lifecycle hooks connect model execution to method-specific logic.
The hooks span prefill and decoding, covering step preparation, operations before and after attention, and the end of \emph{every layer and execution step}.
A \texttt{SparseMethodRuntime}, selected at initialization, implements the callbacks invoked through \texttt{SparseController}.
These callbacks coordinate scoring, selection, and state updates across prompt chunks, decoding steps, and Transformer layers, allowing different sparse methods to share the same model implementation.

During prefill, \texttt{finish\_step} supports H$_2$O's chunk-wise eviction using accumulated attention scores and SnapKV's selection and eviction after the final prompt chunk.
This supports chunked prefill without method-specific model variants such as \texttt{LlamaSnapKV} or \texttt{Qwen3H2O}.
During decoding, \texttt{build\_decode\_selection} performs Quest's query-dependent page selection, while \texttt{finish\_step} supports H$_2$O's cumulative-score updates and online eviction.
Together, \texttt{on\_layer\_end} and \texttt{build\_decode\_selection} also enable OmniKV and DeltaKV to pass observation-layer selection indices to subsequent layers.

\paragraph{Method-Customized \texttt{CacheManager}.}
While hooks specify when method logic runs, \texttt{CacheManager} defines how it stores and updates its persistent state.
Each method controls its KV representation, physical layout, logical-to-physical position mapping, and allocation, write, update, and release operations.
The manager also maintains the persistent metadata associated with its KV state.

Lifecycle callbacks invoke these operations to realize cache changes.
For SnapKV compaction and H$_2$O eviction, the manager updates retained-position mappings, associated method state, and available capacity together.
OmniKV applies reused selection indices to resident KV, whereas KIVI manages quantized KV pages, quantization metadata, and an unquantized residual region.
Crucially, logical selection restricts attention reads without freeing KV; physical eviction additionally updates slot mappings and resident lengths and returns discarded slots to the allocator.
This separation supports diverse access and storage semantics within a shared lifecycle.

\paragraph{Attention Through \texttt{AttentionView}.}
Method-owned storage requires an interface that exposes its contents to attention without imposing a common physical representation.
\texttt{AttentionView} provides this interface.
For a selection-based method,
\begin{equation}
  \mathrm{View}_{\ell,t} = \operatorname{BuildView}(\mathcal{C}_{\ell,t}, \mathcal{I}_{\ell,t}), \quad
  o_{\ell,t} = \operatorname{Attention}(q_{\ell,t}, \mathrm{View}_{\ell,t}),
\label{eq:attention-view}
\end{equation}
where $\ell$ indexes layers, $t$ indexes execution steps, $\mathcal{C}_{\ell,t}$ is the persistent cache state, and $\mathcal{I}_{\ell,t}$ identifies selected tokens or pages.
\texttt{CacheManager} resolves this state and selection into a view describing the accessible data and physical layout for a compatible attention backend.
The underlying state may contain explicit KV tensors or MLA latent and positional states with method-specific metadata.

Views support attention over selected, compressed, quantized, or temporarily reconstructed KV while leaving persistent-state ownership with \texttt{CacheManager}.
For example, Quest views reference selected KV pages; OmniKV views resolve reused indices within each layer's own cache; and DeltaKV views expose KV reconstructed on demand.
Thus, hooks coordinate computation and updates, cache managers implement storage operations, and views supply the resulting payloads to attention.
Appendix~\ref{app:lifecycle-implementation} details view payloads, temporary-buffer lifetimes, and CUDA graph execution.

\paragraph{Execution Overview.}
Figure~\ref{fig:execution-overview} traces inference from input tokens to generated output.
The scheduler processes the prompt in prefill chunks and then feeds each sampled token back for the next decoding step.
It queries \texttt{MemoryOracle} for method-specific capacity and reservation requirements and coordinates execution across GPU ranks (Appendix~\ref{app:scheduling-interfaces}).
Within each Transformer layer, lifecycle hooks run the method-specific selection logic, and \texttt{CacheManager} exposes the required KV through an \texttt{AttentionView}.
Layer-end hooks coordinate state across layers, while the step-end hook applies cache updates before the next forward pass.

\subsection{Extended Cache State Management}
\label{sec:method:cache-state-management}

As agent contexts grow across turns, efficient serving requires both fast execution and reusable persistent state.
SparseEngine extends the lifecycle contract across request to support two complementary operations: continuing from compacted history and selectively pruning cached regions.

\begin{figure}[t]
  \centering
  \includegraphics[width=1.0\textwidth]{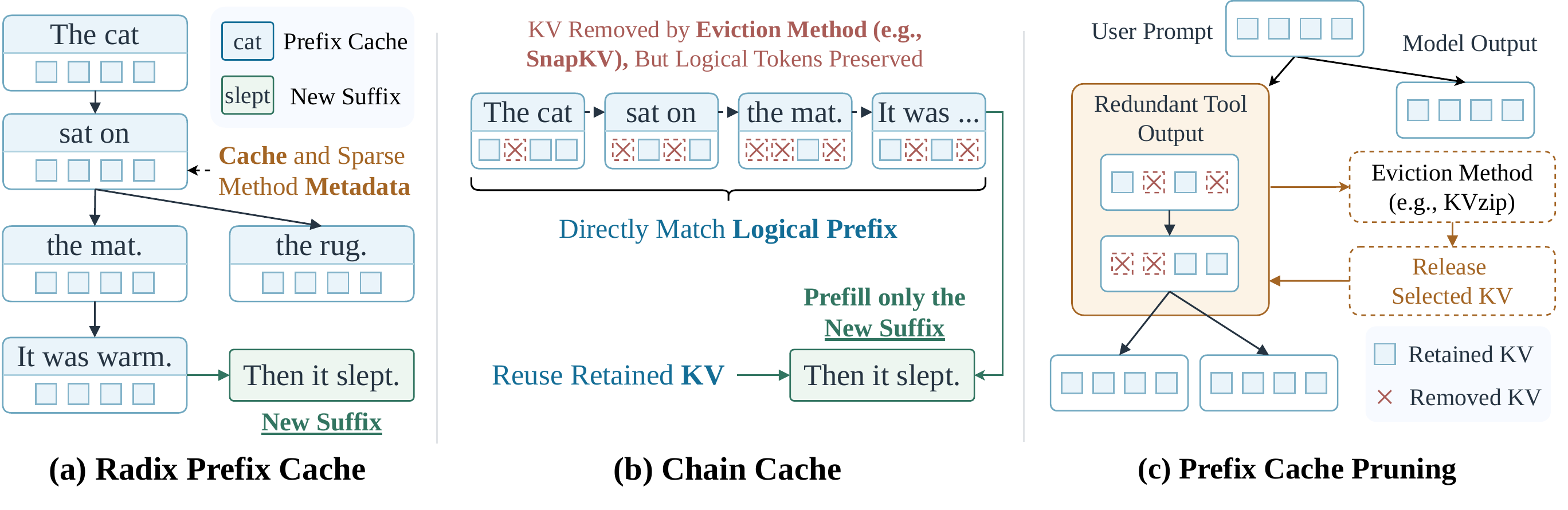}
  \vspace{-2.0em}
  \caption{\textbf{Prefix cache reuse and pruning with sparse methods.}
  (a) Radix prefix caching retains complete KV blocks and method metadata, such as Quest's page min/max summaries.
  (b) After physical eviction, Chain Cache reuses retained KV and method state through logical-prefix matching, prefilling only the unprocessed suffix.
  (c) Prefix Cache Pruning uses an eviction policy to reclaim token-level KV from application-selected history regions while preserving the logical prefix.}
  \label{fig:cache-comparison}
\end{figure}

\paragraph{Chain Cache.}
Multi-turn agents repeatedly extend long trajectories, making prefix caching important for reusing prior computation~\citep{Yao2022ReActSR, Zheng2023SGLangEE, Wadlom2026EfficientLS, Pan2025KVFlowEP}.
Dynamic sparse-attention methods such as Quest, OmniKV, NSA, and DSA retain full KV histories, enabling conventional prefix caching;
eviction-based methods (e.g., SnapKV, KVzip, and H$_2$O) discard entries, so a matching logical prefix no longer implies a complete KV cache.

Chain Cache enables continuation from this compacted state by separating logical-prefix matching from physical KV completeness.
Its \texttt{ChainCacheIndex} associates each session with its logical token prefix, retained per-layer KV, and method-specific metadata.
This state preserves layer-specific retention decisions, as required by SnapKV and H$_2$O, together with persistent statistics such as H$_2$O's accumulated attention scores.
As illustrated in Figure~\ref{fig:cache-comparison}(b), a continuation can therefore reuse the retained state even after physical KV eviction, without reconstructing the complete prefix cache.

Each session receives a \texttt{chain\_id} identifying its state. 
When a request completes, its chain becomes idle while \texttt{CacheManager} retains the per-layer KV and metadata.
A continuation must match both the method configuration and the exact processed logical prefix, although some matched tokens may no longer have resident KV.
The engine then reattaches the retained state and prefills only the unprocessed suffix, carrying previous cache decisions into the next turn.
When capacity is needed, idle chains and their state are reclaimed in least-recently-used order.
Appendix~\ref{app:state-management-details} details continuation boundaries and state-management operations.

\paragraph{Prefix Cache Pruning.}
Chain Cache preserves compacted state for continuation; Prefix Cache Pruning additionally lets applications decide which cached history regions to compact.
Agent trajectories contain user inputs, tool results, reasoning traces, and outputs with different KV utility~\citep{Matam2026MemDecayRK}.
For example, a tool response may become less useful once its relevant information has been incorporated into a reasoning summary.
SparseEngine therefore lets applications specify a history interval and KV retention ratio, then apply a scoring policy such as KVzip~\citep{Kim2025KVzipQK} to select entries within that region.

Concretely, the application supplies the request's complete \texttt{token\_ids} sequence and a block-aligned pruning interval $[L,R)$ with zero-based boundaries, defining $\mathcal{T}={i\mid L\le i<R}$.
A policy such as SnapKV or KVzip returns a mask identifying positions to retain.
The cache manager applies this mask only to idle blocks with no active references or transfers in flight.
Let $P$ denote the logical token prefix and $\mathcal{R}$ the positions with resident KV.
Retaining $\mathcal{I} \subseteq \mathcal{R} \cap \mathcal{T}$ yields:
\begin{equation}
\label{eq:prefix-pruning}
  P'=P, \qquad \mathcal R'=(\mathcal R\setminus\mathcal T)\cup\mathcal I.
\end{equation}
The cache manager releases discarded physical slots and updates resident mappings while preserving $P$ as the prefix-matching key.
Subsequent requests can therefore match the same logical history and reuse its retained KV.
Figure~\ref{fig:cache-comparison}(c) illustrates this operation within a redundant tool-output region. Appendix~\ref{app:state-management-details} details pruning metadata and cache operations.


\section{Experiments}
\label{sec:expt}

\subsection{Experimental Setup}
\label{sec:expt:setup}

We evaluate whether SparseEngine preserves the quality of the original sparse methods on the long-context benchmarks LongBenchV1~\citep{Bai2023LongBenchAB} and LongBenchV2~\citep{Bai2024LongBenchVT}, and assess reasoning quality on AIME~\citep{MaxwellJia2024AIME}.
We further evaluate the impact of sparsity on multi-turn agent task quality using SWE-Bench Lite~\citep{Jimenez2023SWEbenchCL} and Claw-Eval~\citep{Ye2026ClawEvalTT}. We follow the evaluation metrics defined by each benchmark.

For quality evaluation, we compare against the original sparse-method implementations where available to assess whether SparseEngine preserves their task quality, and against full attention to measure the quality impact of sparsity.
For serving performance, we compare SparseEngine against Vortex~\citep{Chen2026VortexEA}, HiSparse~\citep{Xie2026HiSparseSS}, Tangram~\citep{Kim2026TangramUN}, and vLLM (\texttt{v0.26.0})~\citep{Kwon2023EfficientMM}.
Appendix~\ref{app:experimental-settings} gives benchmark protocols and hardware configurations.

For grouped-query attention (GQA), we evaluate dense models represented by Llama-3.1-8B-Instruct~\citep{Dubey2024TheL3} and Qwen3-4B-Thinking-2507~\citep{Yang2025Qwen3TR}, as well as the MoE model Qwen3-30B-A3B-Instruct-2507~\citep{Yang2025Qwen3TR}.
For multi-head latent attention (MLA), we evaluate the MoE model GLM-4.7-Flash~\citep{Zeng2025GLM45AR}.
We also evaluate Qwen3.6-27B~\citep{Yang2025Qwen3TR}, a dense hybrid model that combines linear attention with GQA.
Appendix~\ref{app:supported-models-methods} lists the full set of supported models and sparse methods.
The hardware platforms used across our experiments include NVIDIA H100, H20, RTX PRO 6000, PRO5000, 4090, and 5090 GPUs.

\begin{figure}[t]
  \centering
  \begin{minipage}{0.99\textwidth}
    \centering
    \includegraphics[width=\linewidth]{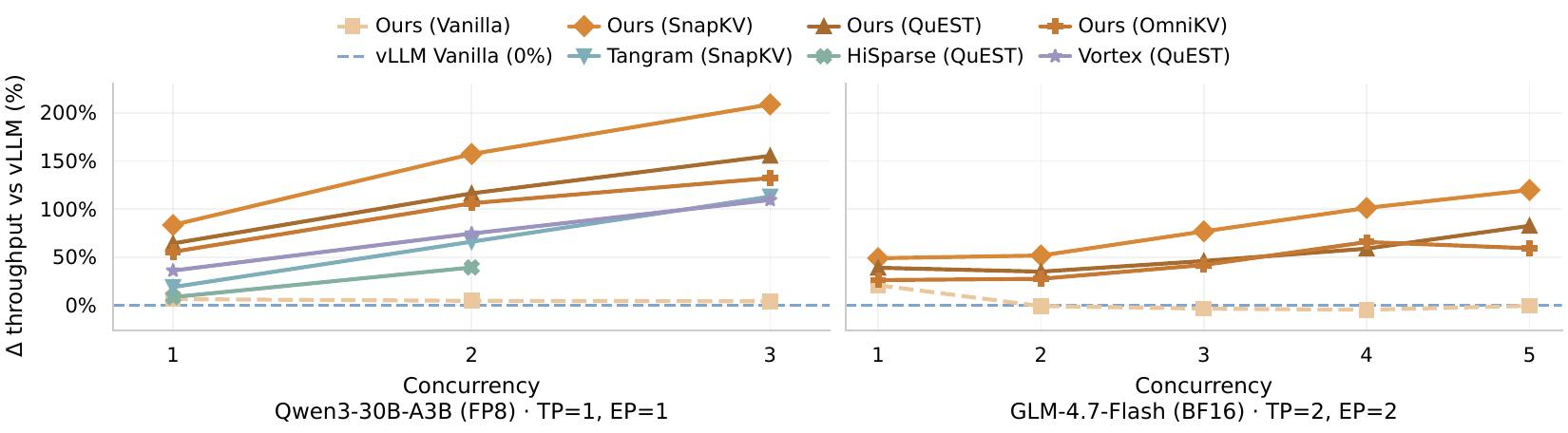}
  \end{minipage}
  \par 
  \begin{minipage}{0.99\textwidth}
    \centering
    \includegraphics[width=\linewidth]{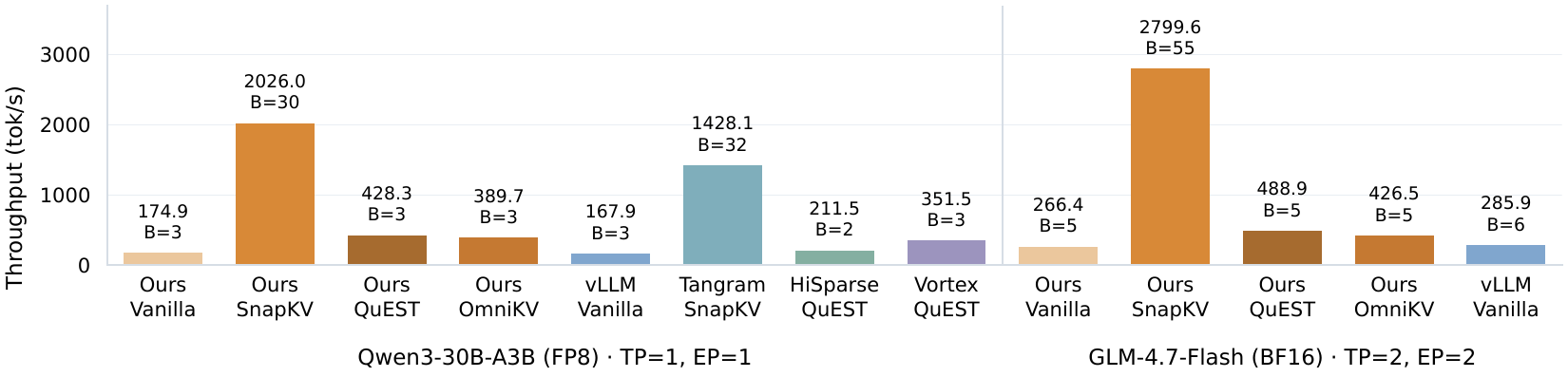}
  \end{minipage}
  \par 
  \begin{minipage}{0.99\textwidth}
    \centering
    \includegraphics[width=\linewidth]{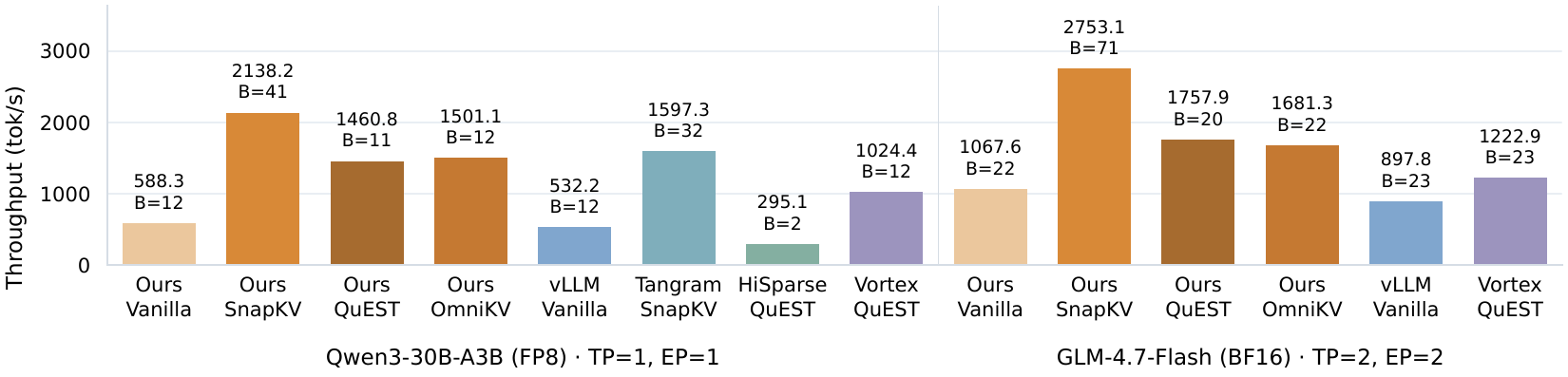}
  \end{minipage}
  \vspace{-0.5em}
  \caption{\textbf{Decode throughput on Qwen3-30B-A3B (left) and GLM-4.7-Flash (right).}
  The upper plot shows throughput improvement over vanilla vLLM at matched batch sizes with 128K-token inputs, with vanilla vLLM defining the 0\% baseline.
  The middle and lower plots show absolute throughput at each system--method pair's largest successfully tested batch size with 128K- and 32K-token inputs, respectively.
  Bar annotations report throughput and batch size ($B$); the largest tested batch size need not be the system's maximum capacity.
  Vortex encounters OOM on GLM with 128K-token inputs even at batch size 1.
  HiSparse and Tangram do not support the evaluated MLA.} 
  \label{fig:decode-performance}
\end{figure}

\subsection{Overall Serving Performance and Quality}
\label{sec:expt:overall}

\begin{table}[t]
  \centering
  \setlength{\tabcolsep}{2pt}
  \renewcommand{\arraystretch}{1.2}
  \definecolor{qualityGain}{HTML}{39734F}
  \definecolor{qualityLoss}{HTML}{A34848}
  \newcommand{\qualityDelta}[2]{{\renewcommand{\arraystretch}{1.1}\tablecell[c]{#1\\[-0.3em]{\fontsize{8}{0.9em}\selectfont\ifdim #2pt>0pt\color{qualityGain}\else\ifdim #2pt<0pt\color{qualityLoss}\else\color{gray}\fi\fi#2}}}}
  \arrayrulecolor{black}
  \caption{\textbf{Sparse-method quality on LongBench V1 and V2.} Deltas below scores report SparseEngine minus the reference implementation in percentage points. Following official evaluation metrics, average score is the six-category average for V1 and accuracy over all examples for V2.}
  \label{tab:quality-longbench}
  \label{tab:quality-longbench-v1}
  \label{tab:quality-longbench-v2}
  \small
  \resizebox{0.99\textwidth}{!}{%
  \begin{tabular}{l ccccccc ccccccc}
    \toprule
    \multirow{2.5}{*}{\textbf{Method}} 
    & \multicolumn{7}{c}{\textbf{LongBenchV1}} & \multicolumn{7}{c}{\textbf{LongBenchV2}} \\
    \cmidrule(lr){2-8} \cmidrule(lr){9-15}
    & \textbf{S-Doc} & \textbf{M-Doc} & \textbf{Summ.} & \textbf{F-Shot} & \textbf{Synth.} & \textbf{Code} & \textbf{Avg.}
    & \textbf{S-Doc} & \textbf{M-Doc} & \textbf{Hist.} & \textbf{Learn.} & \textbf{Struct.} & \textbf{Code} & \textbf{Avg.} \\
    \midrule
    \rowcolor{qualityBlue}
    {\color{qualityInk}\bfseries\emph{GLM-4.7-Flash}\textsuperscript{$\dagger$}} 
    & 41.6 & 47.1 & 27.8 & 71.0 & 52.5 & 65.7 & \textbf{51.0}  
    & 34.3 & 28.8 & 20.5 & 29.6 & 21.2 & 46.0 & \textbf{31.4} \\
    \textbf{SnapKV} 
    & 41.0 & 47.2 & 26.9 & 69.8 & 52.5 & 65.2 & \textbf{50.4} 
    & 36.0 & 31.2 & 20.5 & 32.1 & 30.3 & 42.0 & \textbf{33.2} \\
    \textbf{OmniKV} 
    & 41.7 & 47.1 & 27.8 & 70.6 & 52.5 & 65.1 & \textbf{50.8} 
    & 35.4 & 28.0 & 23.1 & 33.3 & 42.4 & 42.0 & \textbf{33.4} \\
    \textbf{Quest} 
    & 41.4 & 47.2 & 27.7 & 71.2 & 52.5 & 65.4 & \textbf{50.9} 
    & 34.9 & 29.6 & 23.1 & 40.7 & 27.3 & 42.0 & \textbf{33.8} \\
    \textbf{StreamingLLM} 
    & 34.7 & 41.7 & 26.4 & 69.2 & 53.0 & 64.1 & \textbf{48.2} 
    & 34.3 & 32.0 & 23.1 & 34.6 & 30.3 & 32.0 & \textbf{32.4} \\
    \textbf{PyramidKV} 
    & 39.1 & 46.4 & 26.2 & 70.1 & 53.0 & 64.6 & \textbf{49.9} 
    & 35.4 & 26.4 & 25.6 & 34.6 & 24.2 & 36.0 & \textbf{31.6} \\
    \midrule
    \rowcolor{qualityBlue}
    {\color{qualityInk}\bfseries\emph{Llama-3.1-8B}} 
    & \qualityDelta{43.3}{-0.1} & \qualityDelta{46.6}{+0.0} & \qualityDelta{28.8}{+0.0} & \qualityDelta{69.3}{-0.1} & \qualityDelta{55.8}{+0.8} & \qualityDelta{60.2}{+0.7} & \qualityDelta{\textbf{50.7}}{+0.2} 
    & \qualityDelta{33.7}{-1.2} & \qualityDelta{31.2}{+1.6} & \qualityDelta{10.3}{-5.1} & \qualityDelta{25.9}{+2.5} & \qualityDelta{27.3}{-3.0} & \qualityDelta{36.0}{+2.0} & \qualityDelta{\textbf{29.8}}{+0.0} \\
    \addlinespace[0.2em]
    \textbf{SnapKV} 
    & \qualityDelta{43.4}{+0.0} & \qualityDelta{46.6}{+0.0} & \qualityDelta{27.0}{-0.1} & \qualityDelta{68.4}{-0.5} & \qualityDelta{55.9}{+0.8} & \qualityDelta{60.1}{+0.1} & \qualityDelta{\textbf{50.2}}{+0.1} 
    & \qualityDelta{34.3}{+0.6} & \qualityDelta{31.2}{+0.8} & \qualityDelta{10.3}{-2.6} & \qualityDelta{24.7}{+0.0} & \qualityDelta{27.3}{-3.0} & \qualityDelta{36.0}{+0.0} & \qualityDelta{\textbf{29.8}}{+0.0} \\
    \addlinespace[0.2em]
    \textbf{OmniKV} 
    & \qualityDelta{43.2}{-0.2} & \qualityDelta{46.7}{+0.5} & \qualityDelta{29.0}{-0.1} & \qualityDelta{69.1}{+0.0} & \qualityDelta{56.2}{+1.9} & \qualityDelta{60.1}{+0.0} & \qualityDelta{\textbf{50.7}}{+0.4} 
    & \qualityDelta{33.1}{+0.0} & \qualityDelta{30.4}{+0.8} & \qualityDelta{10.3}{-2.6} & \qualityDelta{24.7}{-1.2} & \qualityDelta{27.3}{-3.0} & \qualityDelta{38.0}{+2.0} & \qualityDelta{\textbf{29.4}}{-0.2} \\
    \addlinespace[0.2em]
    \textbf{Quest} 
    & \qualityDelta{43.1}{+0.2} & \qualityDelta{46.1}{-0.2} & \qualityDelta{28.6}{-0.3} & \qualityDelta{68.8}{-0.2} & \qualityDelta{55.4}{+0.5} & \qualityDelta{59.0}{-0.1} & \qualityDelta{\textbf{50.2}}{+0.0} 
    & \qualityDelta{29.1}{-2.3} & \qualityDelta{28.8}{-0.8} & \qualityDelta{12.8}{+7.7} & \qualityDelta{24.7}{+1.2} & \qualityDelta{24.2}{+3.0} & \qualityDelta{32.0}{+0.0} & \qualityDelta{\textbf{27.0}}{+0.0} \\
    \addlinespace[0.2em]
    \textbf{RetroInfer} 
    & \qualityDelta{43.6}{+0.2} & \qualityDelta{46.5}{+0.1} & \qualityDelta{28.9}{-0.1} & \qualityDelta{69.4}{+0.0} & \qualityDelta{55.9}{+0.3} & \qualityDelta{60.5}{+0.5} & \qualityDelta{\textbf{50.8}}{+0.2} 
    & \qualityDelta{32.6}{-0.6} & \qualityDelta{30.4}{+0.0} & \qualityDelta{10.3}{+0.0} & \qualityDelta{24.7}{+1.2} & \qualityDelta{27.3}{-3.0} & \qualityDelta{38.0}{-2.0} & \qualityDelta{\textbf{29.2}}{-0.4} \\
    \addlinespace[0.2em]
    \textbf{StreamingLLM} 
    & \qualityDelta{33.9}{+0.1} & \qualityDelta{40.1}{+0.5} & \qualityDelta{25.5}{+0.1} & \qualityDelta{67.0}{-0.1} & \qualityDelta{52.3}{+0.5} & \qualityDelta{59.4}{+0.3} & \qualityDelta{\textbf{46.4}}{+0.2} 
    & \qualityDelta{33.7}{-0.6} & \qualityDelta{32.8}{+1.6} & \qualityDelta{10.3}{-2.6} & \qualityDelta{19.8}{-4.9} & \qualityDelta{30.3}{+3.0} & \qualityDelta{34.0}{+2.0} & \qualityDelta{\textbf{29.2}}{-0.4} \\
    \addlinespace[0.2em]
    \textbf{PyramidKV} 
    & \qualityDelta{43.4}{-0.1} & \qualityDelta{46.7}{+0.2} & \qualityDelta{27.1}{-0.1} & \qualityDelta{69.0}{+0.0} & \qualityDelta{55.9}{+0.8} & \qualityDelta{59.0}{-0.6} & \qualityDelta{\textbf{50.2}}{+0.0} 
    & \qualityDelta{32.6}{+0.0} & \qualityDelta{28.8}{-2.4} & \qualityDelta{10.3}{-2.6} & \qualityDelta{29.6}{+2.5} & \qualityDelta{33.3}{+3.0} & \qualityDelta{38.0}{+4.0} & \qualityDelta{\textbf{30.0}}{+0.2} \\
    \addlinespace[0.2em]
    \textbf{DeltaKV} 
    & \qualityDelta{43.5}{+0.0} & \qualityDelta{46.2}{+2.0} & \qualityDelta{27.8}{-0.2} & \qualityDelta{69.5}{+1.7} & \qualityDelta{55.6}{+1.8} & \qualityDelta{59.7}{-0.8} & \qualityDelta{\textbf{50.4}}{+0.8} 
    & \qualityDelta{32.6}{+0.6} & \qualityDelta{34.4}{+2.4} & \qualityDelta{10.3}{-5.1} & \qualityDelta{22.2}{-9.9} & \qualityDelta{30.3}{+3.0} & \qualityDelta{32.0}{+2.0} & \qualityDelta{\textbf{29.4}}{-0.8} \\
    \addlinespace[0.2em]
    \textbf{Palu} 
    & \qualityDelta{36.6}{+0.3} & \qualityDelta{36.4}{+0.4} & \qualityDelta{25.2}{+0.0} & \qualityDelta{66.4}{+0.6} & \qualityDelta{47.0}{+0.0} & \qualityDelta{39.0}{-0.2} & \qualityDelta{\textbf{41.8}}{+0.2} 
    & \qualityDelta{32.0}{+1.7} & \qualityDelta{28.8}{+0.0} & \qualityDelta{18.0}{-2.6} & \qualityDelta{28.4}{+2.5} & \qualityDelta{30.3}{+0.0} & \qualityDelta{18.0}{+2.0} & \qualityDelta{\textbf{28.0}}{+1.0} \\
    \addlinespace[0.2em]
    \textbf{TurboQuant} 
    & \qualityDelta{43.5}{+0.3} & \qualityDelta{46.4}{-0.7} & \qualityDelta{29.1}{+0.0} & \qualityDelta{68.8}{+0.0} & \qualityDelta{54.6}{+0.1} & \qualityDelta{59.6}{-0.6} & \qualityDelta{\textbf{50.3}}{-0.1} 
    & \qualityDelta{36.0}{+3.4} & \qualityDelta{27.2}{-6.4} & \qualityDelta{12.8}{+0.0} & \qualityDelta{27.2}{+4.9} & \qualityDelta{39.4}{+12.1} & \qualityDelta{38.0}{+8.0} & \qualityDelta{\textbf{31.0}}{+2.0} \\
    \bottomrule
  \end{tabular}}
  \arrayrulecolor{black}
  \par\vspace{-0.5em}
\end{table}

\paragraph{Long-Context Decode Performance.}
SparseEngine improves long-context decode efficiency.
Physical KV eviction lets SparseEngine sustain substantially larger tested decode batches on the same GPU configuration.
At each system--method pair's largest successfully tested batch size (Figure~\ref{fig:decode-performance}, middle), SparseEngine with SnapKV delivers approximately \textbf{$10\times$} the aggregate decode throughput of vanilla vLLM on both models.
Dynamic sparse attention reduces attention computation by accessing only a selected subset of the KV cache at each decode step, improving throughput at matched batch sizes.
With Quest and OmniKV, SparseEngine reaches roughly $1.5\times$--$2.6\times$ the decode throughput of vanilla vLLM at several matched batch sizes across the two evaluated models (upper).
These results capture complementary benefits of sparse inference: lower attention cost and higher concurrency within the same GPU memory budget.
The lower plot in Figure~\ref{fig:decode-performance} shows that the throughput benefit from larger tested batches also extends to 32K-token inputs.
Appendix~\ref{app:serving-settings} also reports absolute decode throughput for the matched-batch comparison. Appendix~\ref{app:decode-32k} details the 32K settings and complementary results.

\paragraph{Efficiency within Specialized Systems' Supported Methods.}
We further conduct fair, method-specific comparisons against baselines to examine whether SparseEngine's more general implementation comes at the expense of efficiency.
Quest directly matches \textbf{Vortex} and \textbf{HiSparse}'s page-selection abstraction, and SnapKV directly matches \textbf{Tangram}'s KV-retention abstraction~\citep{Chen2026VortexEA,Kim2026TangramUN,Xie2026HiSparseSS}.
At a matched batch size of two on Qwen3 with 128K-token inputs (Figure~\ref{fig:decode-performance}, upper), SparseEngine with Quest delivers $1.24\times$ Vortex's and $1.55\times$ HiSparse's decode throughput.
With SnapKV, SparseEngine delivers $1.55\times$ Tangram's throughput at the same batch size.
SparseEngine remains ahead at every other common measured batch size in the upper plot.
These method-matched results show that SparseEngine's broader method support is compatible with efficient execution even for methods that directly fit the competing systems' abstractions.
Next, we show that our acceleration does not sacrifice the quality of the original sparse methods.

\begin{table}[t]
\centering
\setlength{\tabcolsep}{1.5pt}
\renewcommand{\arraystretch}{1.3}
\arrayrulecolor{black}
\caption{\textbf{AIME 2024 results.} Acc. (\%) averages 60 responses (two per problem). Speedup is relative to Qwen3-4B-Thinking. $^\spadesuit$ denotes the Vortex implementation.}
\label{tab:quality-aime}
\small
\resizebox{0.99\textwidth}{!}{%
\begin{tabular}{lrrr rrr rrrr}
\toprule
\textbf{Method} &
\cellcolor{qualityBlue}\textit{\bfseries Qwen3-4B} &
\textbf{OmniKV} &
\textbf{Quest} &
\textbf{Quest}$^{\spadesuit}$ &
\textbf{SnapKV (L)} &
\textbf{SnapKV (M)} &
\textbf{SnapKV (H)} &
\textbf{PyramidKV} &
\textbf{R-KV} &
\textbf{StreamingLLM} \\
\midrule
\textbf{Acc.} & \cellcolor{qualityBlue}81.7 & 80.0 & 76.7 & 75.0 & 58.3 & 80.0 & 85.0 & 53.3 & 51.7 & 18.3 \\
\textbf{E2E Speedup} & \cellcolor{qualityBlue}$1.00\times$ & $1.51\times$ & $1.28\times$ & $0.91\times$ & $1.66\times$ & $1.33\times$ & $1.05\times$ & $1.66\times$ & $1.30\times$ & $3.36\times$ \\
\bottomrule
\end{tabular}}
\arrayrulecolor{black}
\end{table}

\paragraph{Sparse-Method Quality.}
This experiment evaluates whether sparse methods retain task quality when implemented within SparseEngine.
We evaluate diverse sparse methods on Llama-3.1-8B and GLM-4.7-Flash, using full attention (Vanilla) and the original implementations as references for each model and method.
Table~\ref{tab:quality-longbench} reports the results, with method-specific settings in Appendix~\ref{app:quality-settings}.
Across the 20 overall-score comparisons on LongBench V1 and V2, SparseEngine achieves a mean difference of $+0.17$ points from the reference implementations, with a population variance of $0.32$ squared points.
These small aggregate differences indicate that SparseEngine closely preserves the task quality of the evaluated methods.
Notably, because top-p sampling introduces randomness and the number of instances in LongBench V2 subtasks is relatively small, individual subtask scores show noticeable fluctuations, although the average score remains stable.

\paragraph{End-to-End Reasoning Performance and Efficiency.}
Table~\ref{tab:quality-aime} evaluates task accuracy and end-to-end generation speed on AIME 2024 with Qwen3-4B-Thinking-2507.
In SnapKV, \texttt{Low}, \texttt{Mid}, \texttt{High} represent sparsity budgets of 4k, 8k, and 16k, respectively.
These results also reveal a broad quality-efficiency trade-off: methods offering larger speedups generally suffer greater accuracy drops.
The speedup here is lower than that in Figure~\ref{fig:decode-performance} because the evaluated contexts are shorter. The shorter contexts in this workload reduce attention's share of execution time relative to the long-context decode benchmark, leaving less computation for sparse attention to eliminate.
Appendix~\ref{app:aime-settings} details the evaluation protocol and the method-specific concurrency limits, which reflect different GPU memory requirements.

\subsection{Performance of Cache State Management}
\label{sec:expt:cache}

\begin{figure}[!t]
  \centering
  \captionsetup{skip=0.5em}
  \begin{minipage}[t]{0.59\textwidth}
    \centering
    \includegraphics[height=10.44em]{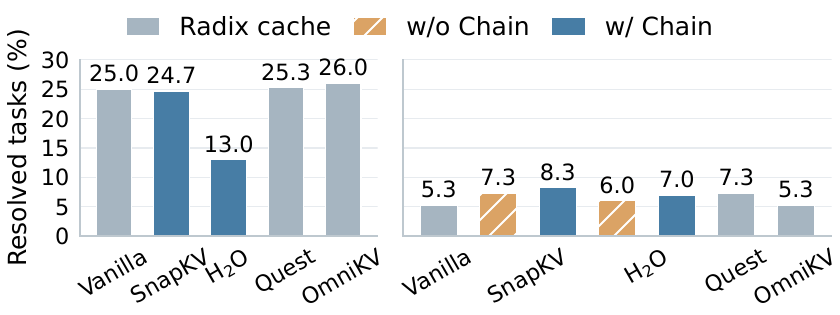}\par
    \caption{ \textbf{SWE-bench Lite resolved tasks (\%) on GLM- 4.7-Flash (left) and Qwen3-30B-A3B (right).} Qwen3 additionally compares Chain Cache on and off.}
    \label{fig:quality-chain-cache}
  \end{minipage}
  \hfill
  \begin{minipage}[t]{0.38\textwidth}
    \centering
    \includegraphics[height=10.44em]{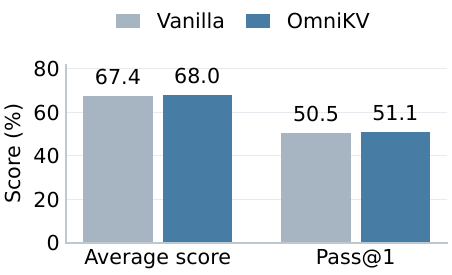}\par
    \caption{\textbf{Claw-Eval quality on Qwen3.6-FP8.} Settings are in Appendix~\ref{app:claw-quality-settings}.}
    \label{fig:quality-claw-eval}
  \end{minipage}
  \vspace{-0.5em}
\end{figure}

\paragraph{Chain Cache Quality on Agent Tasks.}
We evaluate chain cache with mini-SWE-agent on SWE-bench Lite and with Claw-Eval on text-only agent tasks.
SnapKV and H$_2$O retain their compacted KV and method state through Chain Cache, while Quest and OmniKV use radix prefix caching.
Figure~\ref{fig:quality-chain-cache} reports task success alongside a full-attention reference for each model, using the model-specific settings in Appendix~\ref{app:agent-quality-settings}.
SnapKV with Chain Cache achieves a task success rate close to the full-attention reference on GLM and resolves more tasks than the reference on Qwen3, demonstrating multi-turn agent execution with retained compacted state.
On Qwen3, enabling Chain Cache preserves task quality for both SnapKV and H$_2$O compared with re-prefilling the full history at each turn.

\begin{wrapfigure}{R}{0.45\textwidth}
  \vspace{-1.0em}
  \centering
  \includegraphics[width=0.92\linewidth]{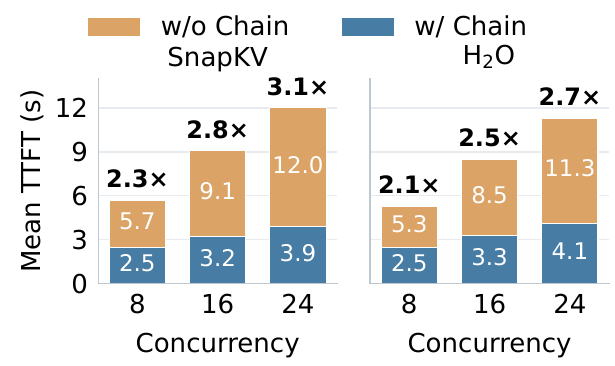}
  \vspace{-1.0em}
  \caption{\textbf{Mean TTFT with and without Chain Cache on Qwen3-30B-A3B-FP8.}}
  \label{fig:chain-cache-ttft}
  \vspace{-1.0em}
\end{wrapfigure}

Figure~\ref{fig:quality-claw-eval} reports Qwen3.6 results on Claw-Eval. OmniKV maintains task quality close to the reference, with settings in Appendix~\ref{app:claw-quality-settings}.
Experiments show that, as long as we don't use an overly aggressive sparsity budget, Chain Cache maintains robust quality on complex multi-turn agent tasks.

\paragraph{Chain Cache Efficiency.}
With the same sparse method and concurrency, enabling Chain Cache reduces mean TTFT by up to $3.1\times$ (Figure~\ref{fig:chain-cache-ttft}).
By reusing retained KV and method metadata across turns, Chain Cache prefills only the newly appended suffix.
It avoids re-prefilling the conversation history and repeating sparse eviction, reducing the work before the first generated token.
Appendix~\ref{app:chain-cache-efficiency-settings} details the replay workload and settings.

\begin{figure}[b]
  \centering
  \arrayrulecolor{black}
  \renewcommand{\arraystretch}{1.2}
  \begin{minipage}[t]{0.59\textwidth}
    \vspace{0em}
    \centering
    \captionof{table}{\textbf{SWE-bench Lite resolved tasks (\%) with prefix pruning.} Pruning uses an 8K trigger and a 2K budget.}
    \label{tab:quality-prefix-pruning}
    \small
    \setlength{\tabcolsep}{3pt}
    \resizebox{0.99\linewidth}{!}{%
    \begin{tabular}{ll cccc}
      \toprule
      \textbf{Decode} & \textbf{Pruning} & \cellcolor{qualityBlue}\textbf{No Pruning} & \textbf{\texttt{lag=0}} & \textbf{\texttt{lag=2}} & \textbf{\texttt{lag=4}} \\
      \midrule
      \textbf{Vanilla} & KVzip & \cellcolor{qualityBlue}25.0 & 24.3 & 25.3 & 25.3 \\
      \textbf{Quest} & KVzip & \cellcolor{qualityBlue}25.3 & 24.3 & 25.7 & 25.3 \\
      \textbf{OmniKV} & SnapKV & \cellcolor{qualityBlue}26.0 & 25.3 & 26.0 & 26.3 \\
      \textbf{OmniKV} & KVzip & \cellcolor{qualityBlue}26.0 & 25.7 & 26.7 & 26.7 \\
      \bottomrule
    \end{tabular}%
    }
  \end{minipage}
  \hfill
  \begin{minipage}[t]{0.38\textwidth}
    \vspace{0.0em}
    \centering
    \includegraphics[width=\linewidth]{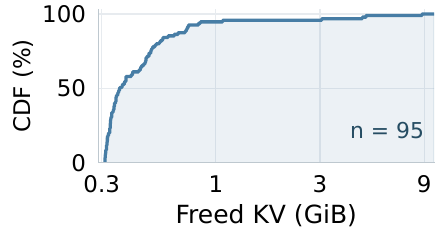}\par
    \vspace{-0.5em}
    \caption{\textbf{KV memory freed per pruning operation.}}
    \label{fig:prefix-pruning-memory}
  \end{minipage}
  \arrayrulecolor{black}
\end{figure}

\begin{figure}[t]
  \centering
  \includegraphics[width=0.99\textwidth]{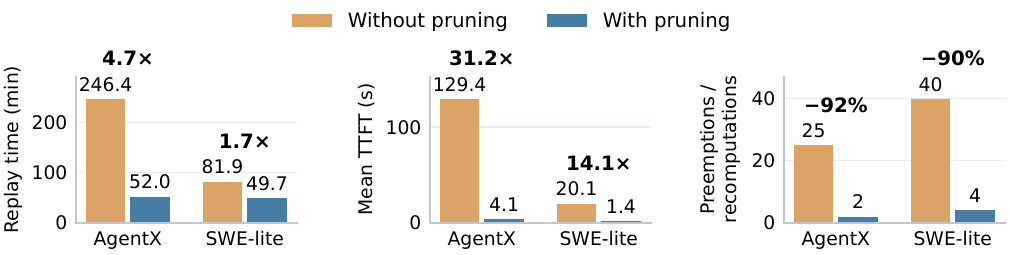}
  \vspace{-0.5em}
  \caption{\textbf{Effect of prefix pruning on agent replay efficiency.} Bold annotations report latency speedup or percentage reduction in preemption/recomputation events. Settings and complete efficiency results are in Appendix~\ref{app:prefix-pruning-quality-settings}.}
  \label{fig:prefix-pruning-efficiency}
  \vspace{-1.0em}
\end{figure}

\paragraph{Prefix Pruning Quality and Efficiency.}
On SWE-bench Lite, delaying tool-result pruning better preserves task performance than immediate pruning, matching or exceeding the unpruned OmniKV baseline with either pruning policy (Table~\ref{tab:quality-prefix-pruning}).
Prefix pruning also accelerates agent replay by up to $4.7\times$ and reduces mean TTFT by up to $31.2\times$ (Figure~\ref{fig:prefix-pruning-efficiency}).
Figure~\ref{fig:prefix-pruning-memory} reports KV memory reclaimed per pruning operation.
Appendix~\ref{app:prefix-pruning-quality-settings} gives the evaluation settings and complete efficiency results.

\paragraph{End-to-End Agent Serving Performance}
Table~\ref{tab:agent-e2e} complements the agent quality evaluation with end-to-end replay of SWE-bench Lite trajectories on GLM-4.7-Flash and a dedicated Gasai agent trace on Qwen3-30B-A3B FP8.
Gasai is used only to evaluate system efficiency.
Replay fixes recorded outputs and includes prefill, decoding, and scheduling, together with simulated tool waits.
All methods use the same GPU configuration, with concurrency selected for each method's KV memory requirements.
SnapKV with an 8K cache budget and Chain Cache reach a \textbf{2.24$\times$} end-to-end speedup over Vanilla on Gasai.
Its memory savings support higher concurrency, while Chain Cache retains compacted state for reuse across turns.
OmniKV also accelerates both workloads at matched concurrency, showing that dynamic sparse attention remains effective with cross-turn prefix reuse.
Together, these results show how SparseEngine translates both sparse computation and KV memory savings into end-to-end serving efficiency.

\begin{table}[t]
\centering
\small
\arrayrulecolor{black}
\setlength{\tabcolsep}{1.5pt}
\renewcommand{\arraystretch}{1.5}
\caption{\textbf{End-to-end agent trace replay performance.} Speedup is relative to each workload's Vanilla at the listed batch sizes. Gasai measures system efficiency only.} 
\label{tab:agent-e2e}
\small
\resizebox{0.99\textwidth}{!}{%
\begin{tabular}{l ccccc cccccc} 
\toprule
\multirow{2.5}{*}{\textbf{Method}} & \multicolumn{5}{c}{\textbf{GLM-4.7-Flash on SWE-lite}} & \multicolumn{6}{c}{\textbf{Qwen3-30B-A3B FP8 on Gasai}} \\
\cmidrule(lr){2-6} \cmidrule(lr){7-12}
& \textbf{Vanilla} & \textbf{Quest} & \textbf{OmniKV} &
\textbf{SnapKV (H)} & \textbf{H$_2$O} &
\textbf{Vanilla} & \textbf{Quest} & \textbf{OmniKV} &
\textbf{SnapKV (M)} &
\textbf{SnapKV (H)} & \textbf{H$_2$O} \\
\midrule
\textbf{Approx. Max BS} & 40 & 32 & 40 & 52 & 80 & 32 & 32 & 32 & 72 & 52 & 80 \\
\textbf{Time (min)} & 98.1 & 92.2 & 62.4 & 82.0 & 48.5 & 49.9 & 58.9 & 30.6 & 22.2 & 30.0 & 25.0 \\
\textbf{E2E Speedup} & $1.00\times$ & $1.06\times$ & $1.57\times$ & $1.20\times$ & $2.02\times$ & $1.00\times$ & $0.85\times$ & $1.63\times$ & $2.24\times$ & $1.66\times$ & $1.99\times$ \\
\textbf{E2E Out. tok/s} & 302.0 & 321.3 & 474.5 & 361.2 & 610.8 & 346.3 & 293.5 & 564.2 & 777.3 & 575.9 & 690.2 \\
\bottomrule
\end{tabular}%
}
\arrayrulecolor{black}
\vspace{0.5em}
\end{table}


\section{Discussions and Limitations}
\label{sec:discussions}


\paragraph{Agent-Assisted Method Integration.}
Frameworks often simplify method integration by standardizing representations or workflows. 
Coding agents could lower the cost of specialized implementations, enabling broader method support while preserving method-specific state and computation.
SparseEngine's lifecycle hooks and method-owned cache state provide an explicit integration boundary for this development model.
An integration harness could combine reference modules, automated checks for method fidelity and cache-lifecycle correctness, and iterative test feedback to accelerate integration while retaining method-specific optimizations and shared serving capabilities.

\paragraph{Serving Natively Sparse Models.}
The same lifecycle abstraction also offers a foundation for serving natively sparse models.
DeepSeek-V3.2 incorporates sparse attention into its architecture~\citep{DeepSeekAI2025DeepSeekV32PT}, while cross-layer index reuse appears in both OmniKV's inference-time sparsification and IndexCache's acceleration of learned sparse attention~\citep{Hao2025OmniKVDC, Bai2026IndexCacheAS}.
SparseEngine's hooks expose when selection state is produced and reused while preserving method-specific cache representations.
For selection-based models retaining full KV histories, eviction could control cache growth, while quantization and compression could reduce the storage cost of retained state.
SparseEngine provides a common foundation for exploring these combinations and extending their state reuse across agent turns.

\paragraph{Limitations of Contract Scope.} SparseEngine's lifecycle contract targets per-request sparse execution and prefix-aligned state management (Chain Cache and Prefix Cache Pruning). Consequently, cross-request paradigms that require non-prefix chunk splicing with mid-sequence recomputation (e.g., CacheBlend~\citep{Yao2025CacheBlendFL}) or cross-prefix and cross-model KV adaptation (e.g., KVComm~\citep{Ye2025KVCOMMOC}) fall outside our current scope and would require additional cross-request abstractions.

\paragraph{Prefill-Decode Disaggregation.}
Prefill-decode (PD) disaggregation is widely used in large-scale production LLM serving to scale prefill and decoding independently~\citep{Zhong2024DistServeDP,Qin2025MooncakeTM}.
SparseEngine does not yet implement PD disaggregation, limiting deployment with separate prefill and decode pools.
Supporting it would require a serialization and transfer protocol for method-owned KV layouts and persistent metadata so that decoding can resume from the transferred state.
Sparse compaction or compression during prefill could reduce the KV data transferred between nodes.
Evaluating these transfer costs remains future work.

\subsection*{AI Use Statement}

We used AI tools to polish the language of the manuscript and assist with implementing portions of the code.
The authors take full responsibility for the manuscript and the accompanying implementation.

\subsection*{Ethics Statement}

This work studies efficient inference and cache management for large language models.
Its intended benefit is to reduce the computational and memory costs of long-context applications.
The proposed system does not address the safety or biases of the underlying models, and deployment should retain appropriate safeguards for the intended application.

\subsection*{Reproducibility Statement}

Appendix~\ref{app:experimental-settings} documents the evaluation protocols, with hardware configurations and method-specific hyperparameters provided in the corresponding subsections.
Appendix~\ref{app:lifecycle-implementation} describes the lifecycle interfaces and cache-management mechanisms used in SparseEngine.
The implementation and reproduction materials are available at \url{https://github.com/CURRENTF/SparseEngine}.

\bibliography{reference}
\bibliographystyle{plainnat}

\appendix
\arrayrulecolor{black}

\section{Detailed Background and Notations}
\label{app:background}

This appendix expands the background introduced in Section~\ref{sec:related-work} with a common notation for attention and cache state, followed by the selection rules of four representative sparse methods.
The equations describe their core mechanisms, with small examples to illustrate how their cache behavior differs.

\subsection{Attention and Cached Representations}
\label{app:background-attention}

\paragraph{Notation.}
Consider one request with a prompt of $n$ tokens.
We use $i$ for a query's token position and $j\leq i$ for a key's position.
These positions differ from the execution-step index $t$ in Section~\ref{sec:method}: a prefill step can process several positions, while a standard decode step processes one position per request.
Table~\ref{tab:background-notation} summarizes the notation.

\begin{table}[ht]
\centering
\renewcommand{\arraystretch}{1.2}
\caption{\textbf{Notation for the background equations.} Layer and head indices are omitted when discussing a single layer and head.}
\label{tab:background-notation}
\small
\begin{tabularx}{0.99\textwidth}{lX}
\toprule
\textbf{Symbol} & \textbf{Description} \\
\midrule
$\ell, a$ & Transformer layer and query-head indices. \\
$n, i, j$ & Prompt length, current query position, and a key position. \\
$[i]$ & The causal history $\{1,\cdots,i\}$, including the current token. \\
$q_{\ell,i}^{(a)}$ & Query vector for head $a$ at position $i$. \\
$k_{\ell,j}^{(a)},v_{\ell,j}^{(a)}$ & Logical key and value used by query head $a$ at position $j$. \\
$z_{i,j},\alpha_{i,j}$ & Query--key logit and normalized attention weight. \\
$\mathcal S_i,\mathcal R_i$ & Positions read by attention at $i$, and positions retained after processing $i$. \\
$\operatorname{TopK}_b(s;\mathcal A)$ & Indices of the $b$ largest scores $s(j)$ among candidates $j\in\mathcal A$, or all candidates if fewer than $b$ exist. \\
$B,\mathcal W_i$ & Retained-token budget and protected recent-token window. \\
\bottomrule
\end{tabularx}
\end{table}

\paragraph{Attention over Selected Positions.}
For one layer and head, write vectors as rows and let $d_k$ be the key dimension.
Given a nonempty set $\mathcal S_i\subseteq[i]$, attention computes:
\begin{equation}
\label{eq:background-selected-attention}
z_{i,j} = \frac{q_i k_j^{\top}}{\sqrt{d_k}}, \quad
\alpha_{i,j}(\mathcal S_i) = \frac{\exp(z_{i,j})}{\sum_{u\in\mathcal S_i}\exp(z_{i,u})}, \quad
o_i = \sum_{j\in\mathcal S_i}\alpha_{i,j}(\mathcal S_i)v_j.
\end{equation}
Dense causal attention uses $\mathcal S_i=[i]$.
Sparse attention restricts this set and normalizes over the selected positions.
Physical eviction additionally changes $\mathcal R_i$, determining which entries remain available to later queries.
For example, reading positions $\{1,4,8\}$ from an eight-token cache can leave all eight entries stored, whereas retaining only these positions prevents later queries from recovering the other five entries from that cache.

\paragraph{MHA, MQA, and GQA.}
Let $H_q$ and $H_{kv}$ be the numbers of query heads and KV heads, and let $g(a)$ map query head $a$ to its KV head.
The logical vectors above satisfy $k_{\ell,j}^{(a)}=k_{\ell,j}^{g(a)}$ and $v_{\ell,j}^{(a)}=v_{\ell,j}^{g(a)}$.
Multi-head attention (MHA) uses a separate KV head for each query head.
Multi-query attention (MQA) shares one KV head across all query heads, while grouped-query attention (GQA) shares each KV head within a group.
For $L$ layers with uniform head dimensions and $N$ cached tokens, explicit KV contains $LN H_{kv}(d_k+d_v)$ scalar elements, excluding metadata.
Sharing KV across heads changes storage cost while preserving the attention computation for each query head.

\paragraph{MLA.}
Multi-head latent attention represents the content keys and values through a shared low-dimensional latent vector.
Suppressing the layer index, let $h_j$ be the hidden state and define:
\begin{equation}
\label{eq:background-mla-latent}
c_j=h_jW^{DKV},\qquad
k_j^{C,(a)}=c_jU_a^K,\qquad
v_j^{(a)}=c_jU_a^V.
\end{equation}
Here $W^{DKV}$ projects to the latent dimension, while $U_a^K$ and $U_a^V$ recover a head's content key and value.
MLA also stores a positional key $k_j^R$, with a corresponding positional query $q_i^{R,(a)}$.
The query's content component is $q_i^{C,(a)}$.
Writing $d_C$ and $d_R$ for their dimensions, the logit and output can be evaluated as:
\begin{equation}
\label{eq:background-mla-absorbed}
z_{i,j}^{(a)} =\frac{(q_i^{C,(a)}(U_a^K)^{\top})c_j^{\top}
       +q_i^{R,(a)}(k_j^R)^{\top}}{\sqrt{d_C+d_R}}, \quad
o_i^{(a)} =\left(\sum_{j\in\mathcal S_i}\alpha_{i,j}^{(a)}(\mathcal S_i)c_j\right)U_a^V.
\end{equation}
This rearrangement computes with cached $(c_j,k_j^R)$ without explicitly storing the full per-head content keys and values.
With latent dimension $d_C$, these tensors contain $LN(d_C+d_R)$ scalar elements under uniform layer dimensions.
Thus, the logical attention rule does not prescribe the physical cache format.
In SparseEngine, the cache manager owns that format and exposes it to a compatible backend through an attention view (Section~\ref{sec:method:architectural-foundations}).

\subsection{Inference Execution and Cache Management}
\label{app:background-engine}

\paragraph{Prefill and Decode.}
Prefill evaluates the prompt with a causal mask, producing the state needed to sample the first output token.
Each later decode step feeds the last sampled token through the model to obtain the next token distribution.
Consequently, the last sampled token has no KV until it is processed in a subsequent forward pass.
Chunked prefill partitions prompt positions into consecutive intervals.
A chunk attends to the preceding cached context and the causally available positions within that chunk.
The scheduler can interleave these chunks with decode work from other requests, while continuous batching updates the participating requests between steps.

\paragraph{Logical Positions and Physical Blocks.}
Paged cache management~\citep{Kwon2023EfficientMM} allocates KV in fixed-capacity physical blocks and maintains a block table for each request.
For blocks of $p$ tokens, an uncompressed token position $j$ has logical block number $\lfloor(j-1)/p\rfloor$ and offset $(j-1)\bmod p$.
The block table resolves this logical block to its physical allocation.
This indirection allows noncontiguous storage and lets attention kernels locate a request's KV.
Sparse compaction additionally needs retained-position mappings because physical offsets no longer need to correspond to consecutive logical tokens.

\paragraph{Reuse and Capacity.}
Prefix caching~\citep{Zheng2023SGLangEE} matches a request's token prefix against previously processed history and reuses compatible cached state.
The scheduler still needs to reserve capacity for the uncached suffix and subsequent generation.
An attention budget measures how much context a computation reads, while a storage budget measures how much state remains allocated.
Quest and OmniKV can read a small subset while retaining the complete history, potentially across memory tiers.
H$_2$O and SnapKV also reduce retained KV through eviction.
This distinction motivates the separate selection and cache-management interfaces in Section~\ref{sec:method:architectural-foundations}.

\subsection{Representative Sparse Methods}
\label{app:background-sparse-methods}

The following equations describe one request.
For H$_2$O, SnapKV, and Quest, layer and head indices are suppressed to emphasize the selection rule.
Each method uses a budget with its own meaning: retained tokens for eviction, or tokens/pages read by a sparse attention operation.

\paragraph{H$_2$O: Accumulated Scores and Online Eviction.}
H$_2$O~\citep{Zhang2023H2OHO} treats tokens receiving substantial accumulated attention as heavy hitters.
Before processing position $i$, the available positions are $\mathcal A_i=\mathcal R_{i-1}\cup\{i\}$.
Attention reads this set and adds its weights to persistent scores:
\begin{equation}
u_i(j)=u_{i-1}(j)+\alpha_{i,j}(\mathcal A_i),\qquad j\in\mathcal A_i,
\label{eq:background-h2o-score}
\end{equation}
where the new token starts with score zero before this update.
For a cache budget $B$ and a protected recent window $\mathcal W_i\subseteq\mathcal A_i$ of at most $B$ positions, the retained set is:
\begin{equation}
\mathcal R_i=\mathcal W_i\cup
\operatorname{TopK}_{B-|\mathcal W_i|}(u_i;\mathcal A_i\setminus\mathcal W_i).
\label{eq:background-h2o-retention}
\end{equation}
KV outside $\mathcal R_i$ is evicted, while the scores of retained entries persist for later updates.
For example, with $B=4$ and two protected recent tokens, the other two slots hold the highest-scoring older tokens.
A recent token that leaves the protected window must compete on its accumulated score to remain cached.
The recurrence explains why continuing H$_2$O requires both retained KV and aligned score state.

\paragraph{SnapKV: Prompt Selection from an Observation Window.}
SnapKV~\citep{Li2024SnapKVLK} estimates which prompt entries will matter during generation by inspecting attention from the prompt's final $w$ positions, $\mathcal O=\{n-w+1,\cdots,n\}$.
For each earlier prompt position, it accumulates attention from these observation queries and pools neighboring scores:
\begin{equation}
u(j)=\sum_{i\in\mathcal O}\alpha_{i,j}([i]),\qquad
s=\operatorname{Pool}_{\kappa}(u),\qquad j\in[n]\setminus\mathcal O.
\label{eq:background-snapkv-score}
\end{equation}
The one-dimensional pooling operator has width $\kappa$ and promotes neighborhoods around important positions.
For example, max pooling sets $s(j)=\max_{r\in\mathcal N_\kappa(j)}u(r)$, where $\mathcal N_\kappa(j)$ is the local prefix neighborhood centered at $j$.
For a prompt-cache budget $B\geq w$, the retained prompt is:
\begin{equation}
\mathcal R_{\mathrm{prompt}}=\mathcal O\cup
\operatorname{TopK}_{B-w}(s;[n]\setminus\mathcal O).
\label{eq:background-snapkv-retention}
\end{equation}
For example, an eight-token prompt with $w=2$ and $B=4$ retains positions 7 and 8 plus two earlier positions selected by the pooled scores.
This selection is computed separately for attention heads, so their retained positions can differ.
The selected prompt KV remains fixed during decoding, while newly generated tokens add their own KV.
The observation window must therefore be available before prompt compaction, including when prefill is chunked.

\paragraph{Quest: Query-Dependent Page Selection.}
Quest~\citep{Tang2024QuestQS} groups token positions into pages $\mathcal P_b$ and stores a minimum and maximum key value for each channel $c$:
\begin{equation}
m_{b,c}=\min_{j\in\mathcal P_b}k_{j,c},\qquad
M_{b,c}=\max_{j\in\mathcal P_b}k_{j,c}.
\label{eq:background-quest-summary}
\end{equation}
For query $q_i$, a page's importance estimate is:
\begin{equation}
\widehat z_{i,b}=\frac{1}{\sqrt{d_k}}
\sum_{c=1}^{d_k}\max\{q_{i,c}m_{b,c},\ q_{i,c}M_{b,c}\}.
\label{eq:background-quest-score}
\end{equation}
This quantity upper-bounds the query-key logit of every token in the page.
It need not equal any individual token's logit because different channels can attain their extrema at different positions.
Quest ranks pages by this estimate and performs attention using the actual KV in the selected pages.
For example, an eight-token history stored in four two-token pages can be searched through four page summaries before loading two selected pages for attention.
Unselected pages remain stored and can be selected by a later query.
New keys update the extrema of their pages, so the page summaries persist alongside the KV history.

\paragraph{OmniKV: Selection Reuse across Layers.}
OmniKV~\citep{Hao2025OmniKVDC} exploits the similarity of important context positions across nearby layers during decoding.
Designated observation (filter) layers inspect the context and produce token indices that subsequent sparse layers reuse.
For illustration, a current-query selector can rank each position by its largest query-key logit across heads:
\begin{equation}
s_{f,i}(j)=\max_a z_{f,i,j}^{(a)},\qquad
\mathcal J_{f,i}=\operatorname{TopK}_b(s_{f,i};[i]),
\label{eq:background-omnikv-selection}
\end{equation}
where $f$ is an observation layer and $b$ is the selected-token budget.
This is a current-query scoring example from the method's \href{https://github.com/antgroup/OmniKV/blob/master/modeling/omnikv.py}{released implementation}.
For a sparse layer $\ell$ assigned to observation layer $f(\ell)$, the defining reuse relationship is:
\begin{equation}
\mathcal I_{\ell,i}=\mathcal J_{f(\ell),i},\qquad
o_{\ell,i}^{(a)}=\sum_{j\in\mathcal I_{\ell,i}}
\alpha_{\ell,i,j}^{(a)}(\mathcal I_{\ell,i})v_{\ell,j}^{(a)}.
\label{eq:background-omnikv-reuse}
\end{equation}
Thus, layers reuse token positions while computing their own attention weights over their own KV.
For example, if an observation layer selects $\{1,4,8\}$, its associated sparse layers each read those positions from their respective caches.
A later decode query can select a different set because historical KV has been retained.
The engine needs to pass selection indices between layers while preserving each layer's cache state.

These mechanisms illustrate the lifecycle requirements used in Section~\ref{sec:method:architectural-foundations}.
H$_2$O updates persistent statistics as inference advances, while SnapKV compacts prompt state after observation-window scoring.
Quest consumes the current query together with persistent page summaries, and OmniKV carries selection results across layers.

\section{Lifecycle Implementation Details}
\label{app:lifecycle-implementation}

\paragraph{Attention View Payloads.}
Separate prefill and decode view types pair a physical payload with request indices, context lengths, and token-slot or page-table coordinates.
Payload types distinguish explicit KV tensors from MLA latent and positional tensors, allowing attention implementations to declare which representations they support.
The cache manager resolves selections into the required physical inputs, including temporary reconstruction when needed.
Reconstruction slots are released after the layer consumes them, while persistent KV remains under the cache manager's ownership.

\paragraph{CUDA Graph Execution.}
For supported CUDA Graph paths, cache managers, method runtimes, and attention implementations provide buffers whose addresses remain stable across replay.
Step preparation updates slot mappings, lengths, and other mutable inputs in these buffers.
The model runner handles capture and replay, while each component keeps its referenced storage alive.
Graph and layout compatibility are checked when the execution path is configured.
This division allows method-specific cache representations and update logic to participate in graph execution through the same lifecycle interfaces.

\subsection{Scheduling Interfaces}
\label{app:scheduling-interfaces}

\paragraph{MemoryOracle.}
The scheduler queries the runtime state on rank 0, whose cache manager provides the capacity reference for admission and step reservations.
A sparse method's attention budget alone does not determine its memory requirement.
Quest may read a small set of pages while retaining a much larger cache.
An eviction method may temporarily hold prompt KV before compaction, and a compressed method may need reconstruction space in addition to its persistent storage.
The scheduler therefore queries method-specific capacity and reservation costs through \texttt{MemoryOracle}, implemented by the runtime state and cache-manager interfaces.

For prompt admission, the interface returns a set of named resource budgets and a request's cost in each budget.
Let $B_j$ be the remaining budget for resource $j$ and $c_j(r)$ the reservation required by request $r$.
The scheduler admits $r$ only when
\begin{equation}
  c_j(r) \leq B_j \quad \text{for every reported resource } j,
  \label{eq:admission}
\end{equation}
and deducts each reservation before considering another request.
The cache manager defines the resources and their units.
For example, PyramidKV can report a separate retained-slot budget for each layer, while DeltaKV reports budgets for full-attention layers, reference entries, and retained raw KV, with space reserved for reconstruction.
This allows methods with heterogeneous storage to expose multiple capacity constraints to the same admission loop.

\paragraph{Step Reservations and Batching.}
Admission reserves capacity for the request, and each execution step also checks the space required for its next prefill chunk or decode token.
The interface reports step costs, available slots, and prefill execution requirements.
These requirements distinguish chunked prefill, full-prompt prefill, and prefill paths that use raw-KV offload.
A compatibility key groups requests that can use the same prefill path.
The scheduler builds a batch within the reported capacity and token limits, reserving space as requests are added.
Requests that cannot fit remain queued or follow the configured failure policy.
When execution requires reclamation or preemption, \texttt{RuntimeState} releases the associated payload through its owner before the capacity is reused.

\subsection{State Reuse and Pruning Details}
\label{app:state-management-details}

\paragraph{Chain Records and Continuation Boundaries.}
A chain record contains an identifier, its active or idle status, a method/configuration fingerprint, and the token count and digest at the processed boundary.
A continuation must match the fingerprint and the exact logical prefix through that boundary before the retained cache can be attached.
The processed boundary excludes the final sampled token when that token has not yet passed through the model, so its KV is computed as part of the continuation.
The driver retains compact logical token IDs to preserve token identity for text-based continuation.
Cancellation, failure, or preemption invalidates the chain and releases its payload and metadata together.
Under tensor parallelism, the driver distributes the admission and victim plan so every rank applies the same state transition.

\paragraph{Pruning Metadata and Cache Operations.}
Applications can inspect radix prefixes, assign retention priorities, and request deletion of eligible subtrees.
Interval pruning accepts a retained-token budget and a scoring policy for a block-aligned history interval.
A zero budget removes all KV entries in that interval.
After pruning, the cache manager records retained offsets within each logical block, and a pruning record marks affected continuations as using compressed history.
Device capacity accounting and subsequent transfers follow the retained slots.
The original logical token history remains the prefix-matching key, while attention uses the reduced physical payload.

\subsection{Sparse-Method Composition}
\label{app:composite-sparse-implementation}

Composition requires coordination between the constituent methods' cache updates and selection state.
The evaluated combinations use method-specific adapters to enforce these dependencies.

\paragraph{Granularity and Index Alignment.}
Quest selects fixed-size pages, whereas SnapKV evicts individual tokens.
For Quest + SnapKV, SnapKV performs physical token compaction at \texttt{finish\_step} during prefill.
The \texttt{CacheManager} then constructs Quest's page summaries, containing the minimum and maximum keys, over the compacted physical slots.

\paragraph{Cross-Layer Consistency.}
OmniKV propagates token indices selected by observation layers to subsequent sparse layers.
For OmniKV + SnapKV, hook configuration enforces a shared cross-layer eviction mask so that retained token positions agree across layers.
Indices propagated through \texttt{on\_layer\_end} therefore map to the corresponding tokens in each layer's physical cache.

\paragraph{Quantized KV Storage.}
The \texttt{CacheManager} stores compacted KV entries in FP8 or TurboQuant layouts and exposes compatible payloads to attention kernels through \texttt{AttentionView}.
This allows quantization to compose with the eviction and dynamic-selection workflows in Tables~\ref{tab:composite-sparse-quality-performance} and~\ref{tab:composite-sparse-turboquant}.

\paragraph{Integration Effort.}
Table~\ref{tab:method-integration-loc} breaks down the method-specific adapter code for the two composite methods described in this subsection.

\begin{table}[t]
\centering
\small
\setlength{\tabcolsep}{2pt}
\renewcommand{\arraystretch}{1.3}
\caption{\textbf{Method-specific effective lines of code for composite methods.} Composite-method adapters are described in Appendix~\ref{app:composite-sparse-implementation}.}
\label{tab:method-integration-loc}
\resizebox{0.99\textwidth}{!}{%
\begin{tabular}{l ccccc}
\toprule
\textbf{Method} & \textbf{Runtime/Algorithm} & \textbf{Cache Manager} & \textbf{Activation Hooks} & \textbf{Method-specific Kernels} & \textbf{Total} \\
\midrule
\textbf{OmniKV + SnapKV} & 81 & 20 & 0 & 0 & 101 \\
\textbf{Quest + SnapKV} & 11 & 125 & 0 & 54 & 190 \\
\bottomrule
\end{tabular}}
\end{table}

\section{Experimental Settings}
\label{app:experimental-settings}

\subsection{Sparse-Method Quality}
\label{app:quality-settings}

\paragraph{Models and Hardware.}
Table~\ref{tab:quality-longbench} evaluates GLM-4.7-Flash and Llama-3.1-8B-Instruct with a maximum model length of 131,072 tokens.
Evaluations run on NVIDIA RTX 4090 and H100 GPUs.
Tensor parallelism is 2 for GLM on LongBench and 1 for all Llama configurations.

\paragraph{Evaluation Protocol.}
LongBench v1 reports category scores and their six-category macro-average, using temperature 0, top-$p$ 1, and top-$k$ 1.
LongBenchV2 uses all examples with the zero-shot direct-answer prompt and the model's chat template.
Inputs exceeding 120,000 tokens undergo middle truncation before chat formatting, with a prompt-token ceiling of 130,944 and at most 128 generated tokens.
Its overall score is \textbf{sample-level accuracy}, including unparsed responses in the denominator.

\paragraph{LongBench Sparse Configurations.}
Table~\ref{tab:longbench-sparse-settings} summarizes method-specific budgets and hyperparameters across LongBench V1 and V2 in Table~\ref{tab:quality-longbench}.
Prefill scoring uses FP32 scores with a chunk size of 8,192 and GPU memory utilization of 0.95.
For H$_2$O, prefill chunk eviction is enabled on both models.

\begin{table}[t]
\centering
\setlength{\tabcolsep}{4pt}
\renewcommand{\arraystretch}{1.4}
\caption{\textbf{Sparse-method configurations for LongBench (V1 and V2) in Table~\ref{tab:quality-longbench}.} Budgets and window sizes are reported in tokens. Baselines follow default hyperparameters from their original papers or official repositories.}
\label{tab:longbench-sparse-settings}
\label{tab:longbench-v2-sparse-settings}
\small
\resizebox{1.0\textwidth}{!}{
\begin{tabular}{l p{4cm} p{4cm} p{4.8cm}}
\toprule
\textbf{Method} & \textbf{LongBench V1 Budget} & \textbf{LongBench V2 Budget} & \textbf{Hyperparameters \& Configurations} \\
\midrule
\textbf{SnapKV}       & 32 recent, 2,016 selected & 0/64 sink, 256 recent, 4,096 selected & 32 observation window, pooling kernel 7 \\
\textbf{Quest}        & 2,048 selected & 64 sink, 256 recent, 4,096 selected & 16-token pages, first two layers dense \\
\textbf{OmniKV}       & 32 recent, 2,048 selected & 64 sink, 256 recent, 4,096 selected & Model-profile full-attention layers \\
\textbf{H$_2$O}       & 2,048 prefill, 2,048 decode & 8,192 prefill, 4,096 decode & 128 prefill scoring window, recent ratio 0.5 \\
\textbf{StreamingLLM} & GLM: 8 / 4,096; Llama: 4 / 2,044 & GLM: 8 / 4,096; Llama: 512 / 4,096 & Retains initial sink tokens and local sliding window \\
\textbf{PyramidKV}    & GLM: 64 / 512 / 4,096; Llama: 8 / 3,978 & 64 sink, 256/512 recent, 4,096 selected & Linear layer ratios 0.6 to 0.01, 32 scoring window \\
\textbf{DeltaKV}      & 8 sink, 128 recent, 2,048 decode & 8 sink, 128 recent, 2,048 decode & Full layers $\{0,2,7,13,16,26\}$, latent dim 512, 4-bit KV \\
\textbf{RetroInfer}   & 4 sink, 64 recent & 4 sink, 64 recent & Retrieval ratio 0.018, estimation ratio 0.232, cluster size 16 \\
\textbf{Palu}         & All tokens & All tokens & Grouped SVD (group size 1), K/V rank 96, BF16 factors \\
\textbf{TurboQuant}   & All tokens & All tokens & 4-bit K/V quantization \\
\bottomrule
\end{tabular}}
\vspace{-1.0em}
\end{table}

\paragraph{Reference Implementations.}
For the corresponding Llama rows in Table~\ref{tab:quality-longbench}, DeltaKV and Palu use the authors' HF implementations, StreamingLLM uses KVCache-Factory, RetroInfer uses the authors' GPU implementation, and TurboQuant uses upstream vLLM 0.20.2 as references.

\subsection{Heterogeneous Sparse-Method Composition}
\label{app:composite-sparse-settings}

Appendix~\ref{app:composite-sparse-implementation} describes the method-specific adapters used in these experiments.

\begin{table}[t]
\centering
\small
\setlength{\tabcolsep}{7pt}
\renewcommand{\arraystretch}{1.2}
\arrayrulecolor{black}
\caption{\textbf{LongBench v1 quality and decoding throughput of heterogeneous sparse-method combinations on PRO5000.} Speedup is relative to Dense with BF16 KV for each model. Settings are in Appendix~\ref{app:composite-sparse-settings}.}
\label{tab:composite-sparse-quality-performance}
\begin{tabular}{ll ccc}
\toprule
\textbf{Method} & \textbf{KV format} & \textbf{LongBench score} & \textbf{Decode (tokens/s)} & \textbf{Speedup} \\
\midrule
\rowcolor{qualityBlue}
\multicolumn{5}{l}{\bfseries\emph{Qwen3-30B-A3B-Instruct-2507-FP8}} \\
\textbf{Dense} & BF16 & 53.2 & 272.0 & 1.0$\times$ \\
\textbf{SnapKV} & BF16 & 53.1 & 640.3 & 2.4$\times$ \\
\textbf{Quest} & BF16 & 53.3 & 639.6 & 2.4$\times$ \\
\textbf{OmniKV} & BF16 & 53.2 & 691.8 & 2.5$\times$ \\
\midrule
\textbf{Quest + SnapKV} & BF16 & 53.4 & 723.5 & 2.7$\times$ \\
\textbf{Quest + SnapKV} & FP8 & 53.1 & 796.6 & 2.9$\times$ \\
\textbf{OmniKV + SnapKV} & BF16 & 53.4 & 897.2 & 3.3$\times$ \\
\textbf{OmniKV + SnapKV} & FP8 & 53.2 & 979.4 & 3.6$\times$ \\
\midrule
\rowcolor{qualityBlue}
\multicolumn{5}{l}{\bfseries\emph{Llama-3.1-8B}} \\
\textbf{Dense} & BF16 & 50.7 & 188.1 & 1.0$\times$ \\
\textbf{Dense} & FP8 & 50.4 & 274.2 & 1.5$\times$ \\
\midrule
\textbf{Quest + SnapKV} & BF16 & 50.2 & 432.7 & 2.3$\times$ \\
\textbf{Quest + SnapKV} & FP8 & 49.7 & 455.4 & 2.4$\times$ \\
\textbf{OmniKV + SnapKV} & BF16 & 50.6 & 458.2 & 2.4$\times$ \\
\textbf{OmniKV + SnapKV} & FP8 & 50.0 & 481.5 & 2.6$\times$ \\
\bottomrule
\end{tabular}
\arrayrulecolor{black}
\end{table}

\begin{table}[!t]
\centering
\small
\setlength{\tabcolsep}{4pt}
\renewcommand{\arraystretch}{1.4}
\caption{\textbf{LongBench v1 quality and decoding throughput with TurboQuant on H100.} The model is Qwen3-30B-A3B-Instruct-2507-FP8. Speedup is relative to BF16 KV within each system--method pair. Input lengths and retained KV budgets are specified in Appendix~\ref{app:composite-sparse-settings}.}
\label{tab:composite-sparse-turboquant}
\resizebox{0.99\textwidth}{!}{%
\begin{tabular}{lll ccc}
\toprule
\textbf{System} & \textbf{Method} & \textbf{KV format} & \textbf{LongBench score} & \textbf{Decode (tokens/s)} & \textbf{Speedup} \\
\midrule
\multirow{2}{*}{\textbf{SparseEngine}}
  & Vanilla & BF16 & 53.2 & 1,069.5 & 1.0$\times$ \\
  & Vanilla & TurboQuant & 53.0 & 323.8 & 0.3$\times$ \\
\midrule
\multirow{2}{*}{\textbf{vLLM}}
  & Vanilla & BF16 & 53.2 & 1,026.2 & 1.0$\times$ \\
  & Vanilla & TurboQuant & 52.2 & 328.2 & 0.3$\times$ \\
\midrule
\multirow{2}{*}{\textbf{SparseEngine}}
  & Quest + SnapKV & BF16 & 53.4 & 1,070.8 & 1.0$\times$ \\
  & Quest + SnapKV & TurboQuant & 53.0 & 533.9 & 0.5$\times$ \\
\midrule
\multirow{2}{*}{\textbf{SparseEngine}}
  & OmniKV + SnapKV & BF16 & 53.4 & 1,224.4 & 1.0$\times$ \\
  & OmniKV + SnapKV & TurboQuant & 52.9 & 549.0 & 0.4$\times$ \\
\bottomrule
\end{tabular}%
}
\end{table}

\paragraph{Models and Hardware.}
Table~\ref{tab:composite-sparse-quality-performance} reports LongBench v1 scores and decoding throughput for Qwen3-30B-A3B-Instruct-2507-FP8 and Llama-3.1-8B on NVIDIA PRO5000 hardware.
The combinations pair Quest or OmniKV with SnapKV and use BF16 or FP8 KV storage.
Table~\ref{tab:composite-sparse-turboquant} evaluates BF16 KV and TurboQuant on NVIDIA H100 hardware using Qwen3-30B-A3B-Instruct-2507-FP8.

\paragraph{PRO5000 Configurations.}
For both models in Table~\ref{tab:composite-sparse-quality-performance}, throughput measurements use a single NVIDIA RTX PRO5000 GPU with tensor-, expert-, and data-parallel sizes of 1.
The batch size is 8, with 32,768 input tokens and a requested output length of 2,048 tokens.

Dense retains the full input history and uses full attention with either BF16 or FP8 KV.
SnapKV retains 8,192 historical KV tokens and attends to all retained KV.
Quest and OmniKV retain the full input history, with a 2,048-token decode attention budget at sparse layers.
Quest + SnapKV and OmniKV + SnapKV retain 8,192 historical KV tokens and use the same 2,048-token attention budget with either BF16 or FP8 KV.

The 2,048-token attention budget includes 64 sink tokens, 512 recent tokens, and 1,472 selected historical tokens.
Quest uses 16-token pages and keeps its first two layers dense.
OmniKV keeps layers $\{0,3,9,18,22,27,43\}$ dense for Qwen3 and $\{0,2,7,13,16,26\}$ for Llama, with zero-based layer indices.

\paragraph{TurboQuant Configurations.}
All throughput measurements in Table~\ref{tab:composite-sparse-turboquant} use a batch size of 8 on a single NVIDIA H100 80GB GPU, with tensor- and expert-parallel sizes of 1.
The composite methods use 32,768 input tokens, retain 8,192 historical KV tokens, and use a 2,048-token decode attention budget at sparse layers.
The standalone Vanilla comparisons in both SparseEngine and vLLM use 8,192 input tokens and retain the full KV history with either BF16 KV or TurboQuant.
We use 8K inputs for Vanilla so that its resident KV length during decoding matches the 8K retained after SnapKV compression.
Input lengths and KV budgets remain unchanged between BF16 and TurboQuant within each method.

\paragraph{Measurement Protocol.}
Both Tables~\ref{tab:composite-sparse-quality-performance} and~\ref{tab:composite-sparse-turboquant} use CUDA Graph execution.
After 32 warmup decoding steps, we measure a contiguous 256-step decoding window with synchronization only at its boundaries, excluding prefill.
Each measurement is repeated three times.

\subsection{AIME Reasoning Performance}
\label{app:aime-settings}

\paragraph{Model and Hardware.}
All methods in Table~\ref{tab:quality-aime} use Qwen3-4B-Thinking-2507 in BF16 on one NVIDIA H100 80GB HBM3 GPU per run, with tensor-, expert-, and data-parallel sizes of 1.

\paragraph{Evaluation Protocol.}
Table~\ref{tab:quality-aime} evaluates all 30 problems in the Maxwell-Jia/AIME\_2024 dataset's train split with two attempts per problem, for 60 generated responses per method.
Accuracy is the fraction of correct responses among all 60 attempts.
Sampling uses temperature 0.6, top-$p$ 0.95, top-$k$ 20, min-$p$ 0, and seed 42, with at most 40,960 generated tokens.

\paragraph{Concurrency and Sparse Configurations.}
Concurrency limits differ across methods to accommodate their GPU memory requirements.
Vanilla and OmniKV use a limit of 24 concurrent requests.
SparseEngine Quest and Vortex Quest both use a limit of 20.
SnapKV, H$_2$O, PyramidKV, R-KV, and StreamingLLM use a concurrency limit of 64.
All runs disable prefix caching and enable CUDA Graph, with a 4,096-token prefill chunk and a 65,536-token batch budget.
Table~\ref{tab:aime-method-settings} summarizes the method-specific budgets and retention settings for Table~\ref{tab:quality-aime}.

\begin{table}[!t]
\centering
\small
\setlength{\tabcolsep}{2pt}
\renewcommand{\arraystretch}{1.5}
\caption{\textbf{Method budgets for the AIME 2024 results in Table~\ref{tab:quality-aime}.} Budgets and windows are in tokens. Selected budgets exclude separately listed sink and recent tokens. }
\label{tab:aime-method-settings}
\resizebox{1.0\textwidth}{!}{
\begin{tabular}{llll p{5.0cm}}
\toprule
\textbf{Method} & \textbf{Sink} & \textbf{Recent} & \textbf{Method Budget} & \textbf{Other Settings} \\
\midrule
\textbf{OmniKV}               & 16 & 64 & 2,048 selected & Model-profile full-attention layers \\
\textbf{Quest (SparseEngine)} & 16 & 64 & 2,992 selected (3,072 total) & 16-token pages, first two layers dense \\
\textbf{Quest (Vortex)}       & 16 & 64 & 2,048 selected (128 pages; 2,128 total) & First two layers dense \\
\textbf{SnapKV}               & 16 & 64 & 4,096 selected (4,176 total) & 16 observation window, eviction every 1,024 decode steps \\
\textbf{H$_2$O}               & -- & -- & 4,096 prefill, 4,096 decode & 128 prefill scoring window, recent ratio 0.2, eviction every 1,024 decode steps \\
\textbf{PyramidKV}            & 16 & 64 & 7,987 configured selected & 32 observation window, linear layer ratios, eviction every 1,024 decode steps \\
\textbf{R-KV}                 & 16 & 64 & 4,096 selected & Compression interval 1,024, 8 observation tokens, $\alpha=0.1$, pooling kernel 7 \\
\textbf{StreamingLLM}         & 64 & 2,048 & -- & -- \\
\bottomrule
\end{tabular}}
\vspace{-1.0em}
\end{table}

\subsection{Multi-Turn Agent Quality}
\label{app:agent-quality-settings}

\paragraph{Evaluation Protocol.}
Figure~\ref{fig:quality-chain-cache} reports GLM-4.7-Flash in BF16 and Qwen3-30B-A3B-Instruct-2507-FP8. We use mini-SWE-agent with the canonical 300 test tasks of SWE-bench Lite and the official Docker evaluator.
GLM-4.7-Flash runs on two NVIDIA H100 80GB GPUs with tensor and expert parallelism of 2.
Qwen3-30B-A3B-Instruct-2507-FP8 runs on one NVIDIA H100 GPU with tensor parallelism of 1.
Thinking is enabled and retained across turns.
The context limit is 202K tokens, with at most 16,384 generated tokens per response, temperature 0.7, top-$p$ 0.8, and top-$k$ 20.
Table~\ref{tab:agent-quality-cache-settings} summarizes the GLM and Qwen3 cache settings for Figure~\ref{fig:quality-chain-cache}.

The Qwen3 Chain Cache ablation in Figure~\ref{fig:quality-chain-cache} evaluates SnapKV and H$_2$O with and without cross-turn reuse on NVIDIA H100 hardware using Qwen3-30B-A3B-Instruct-2507-FP8.
Within each method, the sparse configuration is held fixed.
Without Chain Cache, each turn re-prefills the conversation history and re-executes sparse eviction.
With Chain Cache, retained KV entries and method metadata are reused across turns.

\begin{table}[!t]
\centering
\small
\setlength{\tabcolsep}{3pt}
\renewcommand{\arraystretch}{1.5}
\arrayrulecolor{black}
\caption{\textbf{Cache settings for the agent-quality results in Figure~\ref{fig:quality-chain-cache}.} Budgets are in tokens and retain each method's configured meaning.}
\label{tab:agent-quality-cache-settings}
\resizebox{0.99\textwidth}{!}{%
\begin{tabular}{lllll p{6cm}}
\toprule
\textbf{Method} & \textbf{Sink} & \textbf{Recent} & \textbf{Method Budget} & \textbf{Prefix Cache} & \textbf{Other Settings} \\
\midrule
\rowcolor{qualityBlue}
\multicolumn{6}{l}{\bfseries\emph{GLM-4.7-Flash}} \\
\textbf{SnapKV} & 64 & 512 & 15,808 selected & Chain & 32 scoring window; probability-based prefill scores; no full-attention layers \\
\textbf{H$_2$O} & -- & -- & 16,384 prefill; 8,192 decode & Chain & 128 prefill scoring window; recent ratio 0.5; FP32 logit scoring; decode eviction disabled \\
\textbf{Quest} & 64 & 512 & 1,472 selected (2,048 total) & Radix & 16-token pages; first two layers dense \\
\textbf{OmniKV} & 64 & 512 & 1,472 selected (2,048 total) & Radix & Model-profile full-attention layers \\
\midrule
\rowcolor{qualityBlue}
\multicolumn{6}{l}{\bfseries\emph{Qwen3-30B-A3B}} \\
\textbf{Vanilla} & -- & -- & Full KV & Radix & -- \\
\textbf{SnapKV} & 0 & 32 & 8,192 selected & Chain & 32 scoring window; pooling kernel 7 \\
\textbf{Quest} & 0 & 32 & 8,192 selection budget & Radix & 16-token pages; no full-attention layers \\
\textbf{OmniKV} & 0 & 32 & 8,192 selected & Radix & Full-attention layers $\{0,3,9,18,22,27,43\}$ \\
\bottomrule
\end{tabular}}
\arrayrulecolor{black}
\end{table}

\newpage
\subsection{Chain Cache Efficiency}
\label{app:chain-cache-efficiency-settings}

\begin{wrapfigure}{r}{0.4\textwidth}
  \vspace{-1.5em}
  \centering
  \includegraphics[width=0.95\linewidth]{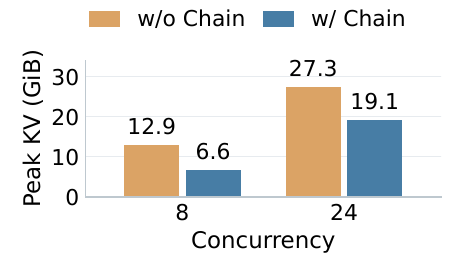}
  \vspace{-1.0em}
  \caption{\textbf{SnapKV peak KV memory.} Settings are specified in this subsection.}
  \label{fig:chain-cache-memory}
  \vspace{-1.0em}
\end{wrapfigure}

Figure~\ref{fig:chain-cache-ttft} and Figure~\ref{fig:chain-cache-memory} use Qwen3-30B-A3B-Instruct-2507-FP8 on NVIDIA H100 hardware.
We replay 32 multi-turn SWE-agent trajectories with 415K total output tokens.
The TTFT experiment evaluates SnapKV and H$_2$O at concurrencies 8, 16, and 24.
Peak occupied KV memory is measured for SnapKV at concurrencies 8 and 24.
Within each pair, the sparse method and concurrency are held fixed.
Without Chain Cache, each turn re-prefills the conversation history and re-executes sparse eviction.
With Chain Cache, retained KV entries and method metadata are reused across turns.

\subsection{Tool-Result Prefix Pruning Quality}
\label{app:prefix-pruning-quality-settings}

Table~\ref{tab:quality-prefix-pruning} evaluates GLM-4.7-Flash-FP8 on NVIDIA H100 hardware with the same mini-SWE-agent task set and evaluator as Appendix~\ref{app:agent-quality-settings}.
Each configuration runs a closed-loop evaluation with its own tool trajectories.
Hardware and precision are matched across no-pruning baselines and all pruning configurations.

Pruning targets tool-result bodies, triggering when cumulative tool tokens reach 8K and compacting to a 2K-token budget.
The experiment compares SnapKV and KVzip pruning policies combined with Vanilla, Quest, or OmniKV decoding, as shown in Table~\ref{tab:quality-prefix-pruning}.
The no-pruning baselines retain the tool-result KV.
Immediate pruning uses \texttt{lag=0}; delayed pruning uses \texttt{lag=2} or \texttt{lag=4}, leaving the latest two or four tool-result rounds intact, respectively.

Figure~\ref{fig:prefix-pruning-efficiency} and Table~\ref{tab:prefix-pruning-efficiency} evaluate replay efficiency with GLM-4.7-Flash-FP8 on NVIDIA H100 hardware.
AgentX (a multi-turn agent trace) contains 20 sessions with 1,108 requests and 1.1M output tokens, replayed at concurrency 20.
SWE-bench Lite contains 64 trajectories with 4,253 requests, replayed at concurrency 64.
Figure~\ref{fig:prefix-pruning-memory} shows the distribution of KV memory released per pruning operation.

\begin{table}[t]
\centering
\small
\setlength{\tabcolsep}{10pt}
\renewcommand{\arraystretch}{1.2}
\caption{\textbf{Prefix pruning efficiency on agent replay.} w/o and w/ denote pruning disabled and enabled. Replay settings are in Appendix~\ref{app:prefix-pruning-quality-settings}. Figure~\ref{fig:prefix-pruning-efficiency} visualizes replay time and mean TTFT.}
\label{tab:prefix-pruning-efficiency}
\begin{tabular}{lcccc}
\toprule
\multirow{2.5}{*}{\textbf{Metric}}
& \multicolumn{2}{c}{\textbf{AgentX}} & \multicolumn{2}{c}{\textbf{SWE-bench Lite}} \\
\cmidrule(lr){2-3}\cmidrule(lr){4-5}
& \textbf{w/o} & \textbf{w/} & \textbf{w/o} & \textbf{w/} \\
\midrule
\textbf{Replay time (min)} & 246.4 & 52.0 & 81.9 & 49.7 \\
\textbf{Throughput (tokens/s)} & 79.1 & 374.6 & 204.3 & 336.7 \\
\textbf{Mean TTFT (s)} & 129.4 & 4.1 & 20.1 & 1.4 \\
\textbf{Mean request E2E latency (s)} & 200.8 & 28.8 & 36.2 & 13.4 \\
\textbf{Preemptions} & 25 & 2 & 40 & 4 \\
\bottomrule
\end{tabular}
\end{table}

\subsection{Claw-Eval Agent Quality}

\label{app:claw-quality-settings}

Figure~\ref{fig:quality-claw-eval} evaluates the 182 text-only tasks among the 199 general Claw-Eval tasks, with one trial per task and DeepSeek-V4-Flash as the judge. Qwen3.6-27B-FP8 runs on one NVIDIA H100 80GB HBM3 GPU with tensor-parallel size 1, eight concurrent workers, a 65,536-token context limit, and at most 4,096 generated tokens per response.
Both methods use prefix caching, a 4,096-token prefill chunk. GPU memory utilization is 0.9. OmniKV retains 2,048 selected tokens and 32 recent tokens, with no sink tokens, and keeps layers $\{3,19,35,51\}$ dense.
Mean scores are reported as percentages over valid tasks. Pass@1 uses all 182 tasks for both methods.

\begin{figure}[!t]
\centering
\includegraphics[width=0.99\textwidth]{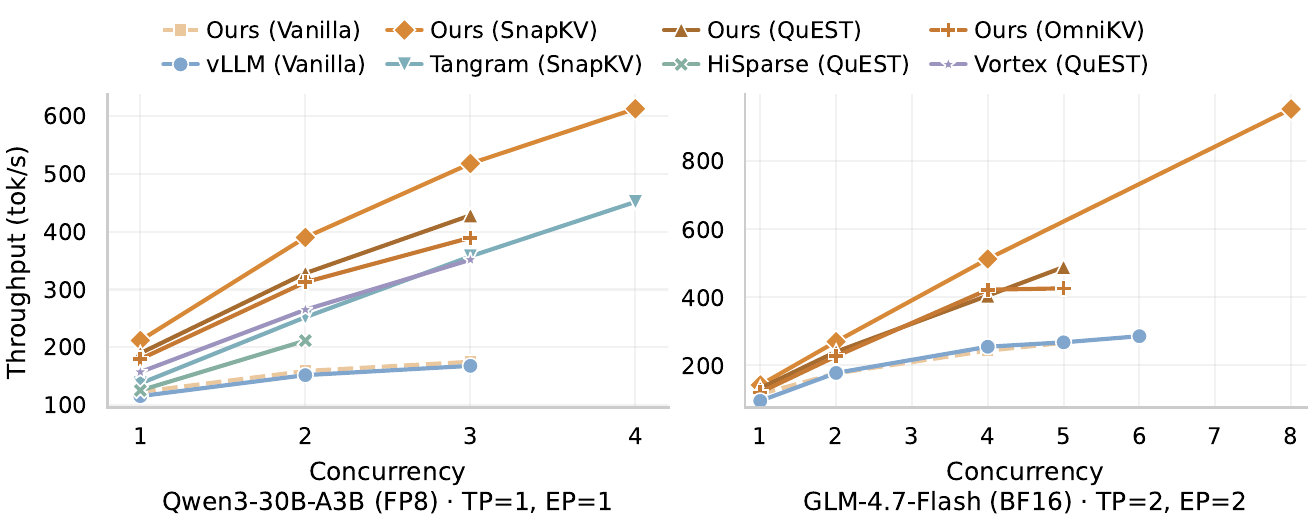}
\vspace{-0.5em}
\caption{\textbf{Absolute decode throughput.} Under 128K-token inputs and 2K-token outputs on Qwen3-30B-A3B (left) and GLM-4.7-Flash (right). Ours denotes SparseEngine. Vortex encounters OOM on GLM at batch size 1, and HiSparse and Tangram do not support the evaluated GLM MLA configuration.}
\label{fig:decode-128k-absolute}
\end{figure}

\subsection{Overall Serving Performance}
\label{app:serving-settings}

\paragraph{Long-Context Decode Benchmark.}
Figures~\ref{fig:decode-performance} and~\ref{fig:decode-128k-absolute} evaluate Qwen3-30B-A3B-Instruct-2507-FP8 on one NVIDIA H100 80GB HBM3 GPU and GLM-4.7-Flash in BF16 on two such GPUs.
The tensor- and expert-parallel sizes are both 1 for Qwen3 and both 2 for GLM.
Each request uses 131,072 input tokens and 2,048 output tokens, with GPU memory utilization set to 0.9.

For both models in Figure~\ref{fig:decode-performance}, SparseEngine allocates 64 sink and 512 recent tokens.
Decode retention is set to 7,616 tokens for SnapKV and 1,472 tokens for Quest and OmniKV, with OmniKV retaining its model-profile full-attention layers.
SnapKV uses an observation window of 32 tokens with probability-based prefill scoring.
For the 128K SnapKV runs, requests are admitted in waves of two with one decode step between waves, using an 8,192-token prefill chunk and batch-token budget. Measurement begins only after the full batch is resident.

\paragraph{External Sparse Baselines.}
Vortex uses Quest with 16-token pages: 4 sink pages, 32 recent pages, and 92 selected pages, with layers 0 and 1 kept dense.
HiSparse uses Quest with 16-token pages, 32 recent pages, and a sparsity ratio of $1/85$.
Tangram uses SnapKV with a uniform 8,192-token compression budget, 64 sink tokens, a 512-token recent window, a scoring window of 32, a pooling kernel of 7, and a prefill chunk size of 4,096.

\paragraph{Measurement Protocol.}
After one discarded workload, we measure three repetitions, each with 32 warmup decode steps followed by a contiguous 256-step full-batch decode window.
Decode throughput is the total number of decoded tokens across these windows divided by their total elapsed time, excluding prefill. GPU synchronization occurs only at window boundaries.
Each request completes the full 2,048-token output, and a batch size is accepted only when all requests remain resident throughout the measured window and the run completes successfully.
The upper plot in Figure~\ref{fig:decode-performance} reports $100(T/T_{\mathrm{vLLM}}-1)$ at matched batch sizes, where $T$ denotes decode throughput.
Relative throughput uses only exactly matched measured batch sizes, without interpolation.
Figure~\ref{fig:decode-128k-absolute} gives the absolute throughput corresponding to the low-batch comparison in the upper plot of Figure~\ref{fig:decode-performance}.

\subsection{Method-Matched Decoding Latency}
\label{app:specialized-latency-settings}

\paragraph{Experimental Setup.}
Tables~\ref{tab:specialized-decode-latency} and~\ref{tab:specialized-decode-breakdown} evaluate Qwen3-30B-A3B-Instruct-2507-FP8 with FP8 weights and BF16 KV on the same NVIDIA H100 80GB HBM3 GPU.
Tensor-, expert-, and data-parallel sizes are all 1.
Each request uses 131,072 input tokens and 2,048 output tokens, with CUDA Graph execution enabled and GPU memory utilization set to 0.9.
Table~\ref{tab:specialized-decode-latency} compares Quest against Vortex and HiSparse at concurrency 2, and SnapKV against Tangram at concurrency 16.
The comparisons use matched sparse budgets and configurations.
Table~\ref{tab:specialized-decode-breakdown} reports component timings for the same system-method comparisons.

\paragraph{Measurement Protocol.}
Table~\ref{tab:specialized-decode-latency} reports decoding latency measured without profiling.
Each measurement uses 32 warmup decoding steps followed by a contiguous 256-step decoding window and is repeated three times.
Table~\ref{tab:specialized-decode-breakdown} reports component-level timings from separate Nsight Systems profiling runs.

\begin{table}[!htbp]
  \centering
  \small
  \renewcommand{\arraystretch}{1.4}
  \caption{\textbf{Method-matched decoding latency against specialized sparse-serving systems.} Latencies are in ms/step and are measured without profiling. Settings are in Appendix~\ref{app:specialized-latency-settings}.}
  \label{tab:specialized-decode-latency}
  {
  \resizebox{0.8\textwidth}{!}{%
  \begin{tabular}{l c l ccc}
    \toprule
    \textbf{Method} & \textbf{Concurrency} & \textbf{Baseline} & \textbf{SparseEngine} & \textbf{Baseline} & \textbf{Speedup} \\
    \midrule
    \textbf{Quest} & 2 & Vortex & 5.8 & 7.2 & 1.2$\times$ \\
    \textbf{Quest} & 2 & HiSparse & 5.8 & 9.4 & 1.6$\times$ \\
    \textbf{SnapKV} & 16 & Tangram & 10.0 & 15.0 & 1.5$\times$ \\
    \bottomrule
  \end{tabular}%
  }\par}
\par\vspace{1em}
  \setlength{\tabcolsep}{5pt}
  \caption{\textbf{Decoding latency breakdown for matched sparse methods.} Component timings are in ms/step and come from separate profiling runs. Settings are in Appendix~\ref{app:specialized-latency-settings}.}
  \label{tab:specialized-decode-breakdown}
  \resizebox{0.99\textwidth}{!}{%
  \begin{tabular}{ll ccccc}
    \toprule
    \multirow{2}{*}[2.5ex]{\textbf{System}} & \multirow{2}{*}[2.5ex]{\textbf{Method}} & \multirow{2}{*}[2.5ex]{\textbf{Concurrency}} & \multirow{2}{*}[2.5ex]{\textbf{Attention}} & \textbf{\shortstack{Sparse\\Selection}} & \textbf{\shortstack{KV Layout/\\Metadata}} & \textbf{\shortstack{Other\\Compute}} \\
    \midrule
    \textbf{SparseEngine} & Quest & 2 & 0.95 & 1.00 & 0.16 & 3.68 \\
    \textbf{Vortex} & Quest & 2 & 1.27 & 1.06 & 0.13 & 4.22 \\
    \textbf{HiSparse} & Quest & 2 & 0.66 & 4.28 & 0.24 & 3.84 \\
    \midrule
    \textbf{SparseEngine} & SnapKV & 16 & 5.02 & -- & 0.07 & 3.83 \\
    \textbf{Tangram} & SnapKV & 16 & 5.13 & -- & 0.12 & 9.24 \\
    \bottomrule
  \end{tabular}%
  }
\end{table}

\subsection{Serving Performance with 32K-token Inputs}
\label{app:decode-32k}

\begin{figure}[!t]
  \centering
  \includegraphics[width=0.99\textwidth]{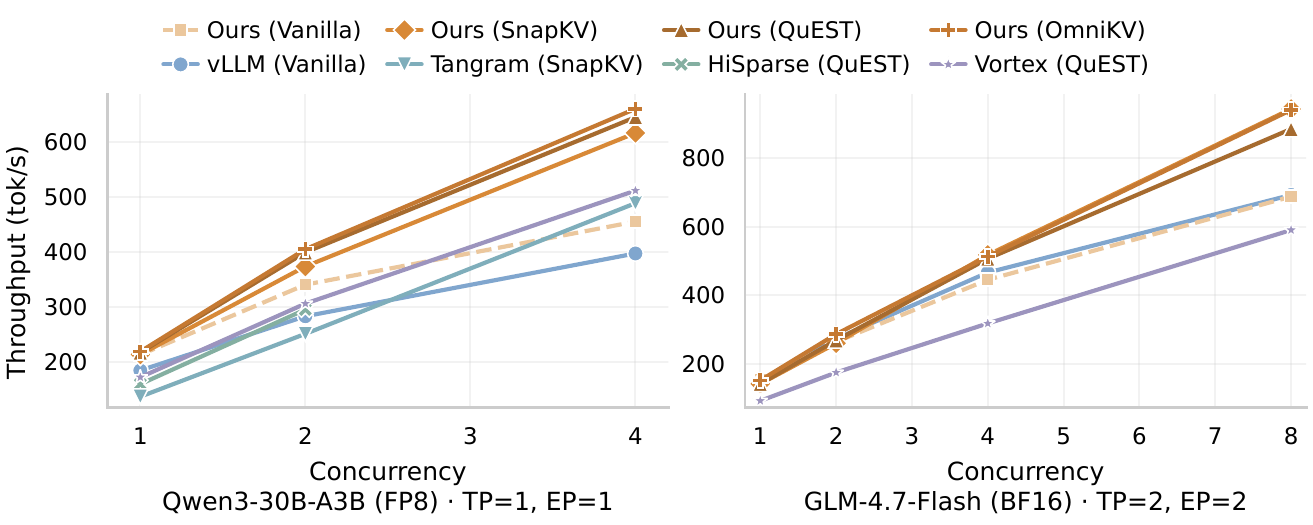}
  \vspace{-0.5em}
  \caption{\textbf{Absolute decode throughput.} Under 32K-token inputs and 2K-token outputs on Qwen3-30B-A3B (left) and GLM-4.7-Flash (right). Ours denotes SparseEngine. HiSparse and Tangram are omitted for GLM because they do not support the evaluated MLA configuration.}
  \label{fig:decode-32k-absolute}
  \vspace{-0.5em}
\end{figure}

\begin{figure}[!t]
  \centering
  \includegraphics[width=0.99\textwidth]{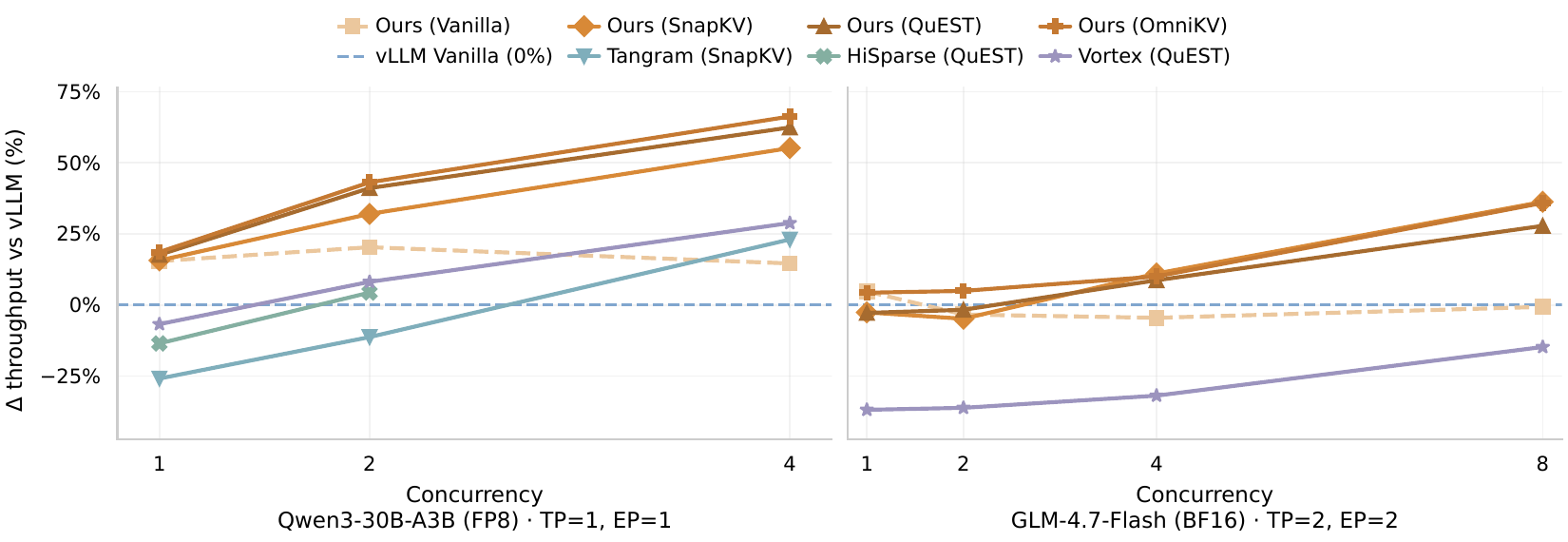}
  \vspace{-0.5em}
  \caption{\textbf{Decode throughput improvement over vanilla vLLM at matched batch sizes with 32K-token inputs and 2K-token outputs.} Ours denotes SparseEngine; vanilla vLLM defines the 0\% baseline.
  HiSparse and Tangram do not support the evaluated GLM MLA configuration and are omitted for that model.}
  \label{fig:decode-32k-capacity}
  \vspace{-0.5em}
\end{figure}

\paragraph{Experiments Setup.}
The lower plot in Figure~\ref{fig:decode-performance} and Figures~\ref{fig:decode-32k-absolute} and~\ref{fig:decode-32k-capacity} use 32,768 input tokens and 2,048 output tokens per request.
The models, hardware, GPU memory utilization, and measurement protocol follow Appendix~\ref{app:serving-settings}.

SparseEngine uses the same sink, recent, and decode budgets as the 128K benchmark, alongside the SnapKV scoring settings above.
External baseline settings are unchanged, except that HiSparse uses a sparsity ratio of $1/21$ and Tangram uses a pooling kernel of 1.

\paragraph{Results.}
SparseEngine maintains higher Quest decode throughput than Vortex at the matched batch sizes shown for both models in Figure~\ref{fig:decode-32k-capacity}.
The SnapKV results in the lower plot of Figure~\ref{fig:decode-performance} further demonstrate how physical KV eviction enables higher serving concurrency within the same GPU memory budget.
Together with the 128K results, these experiments support efficient execution across different context lengths under SparseEngine's shared lifecycle contract.

\subsection{Supported Models and Sparse Methods}
\label{app:supported-models-methods}

Table~\ref{tab:supported-models} lists 14 supported model variants spanning grouped-query attention, multi-head latent attention, hybrid linear attention, and sliding-window/global attention, with both dense and mixture-of-experts FFNs.
Table~\ref{tab:supported-sparse-methods} groups 15 cache methods and FlashPrefill-v2 into the four categories introduced in Figure~\ref{fig:system-comparison}.

These integrations cover different attention paths, KV representations, and state-update workflows, demonstrating that SparseEngine's shared lifecycle contract can accommodate diverse model architectures and sparse inference mechanisms.

\begin{table}[t]
\centering
\small
\renewcommand{\arraystretch}{1.4}
\caption{\textbf{Supported models and their attention and FFN architectures.} GQA denotes grouped-query attention, MLA denotes multi-head latent attention, and MoE denotes mixture of experts. Hybrid linear attention combines GQA with Gated DeltaNet layers.}
\label{tab:supported-models}
\begin{tabularx}{0.99\textwidth}{llXl}
\toprule
\textbf{No.} & \textbf{Model} & \textbf{Attention Architecture} & \textbf{FFN Architecture} \\
\midrule
1 & Qwen2.5~\citep{Yang2024Qwen25TR} & GQA & Dense \\
2 & Qwen3 Dense~\citep{Yang2025Qwen3TR} & GQA & Dense \\
3 & Qwen3 MoE~\citep{Yang2025Qwen3TR} & GQA & MoE \\
4 & Qwen3.5 Dense & Hybrid linear attention & Dense \\
5 & Qwen3.5 MoE & Hybrid linear attention & MoE \\
6 & Qwen3.6 Dense & Hybrid linear attention & Dense \\
7 & Qwen3.6 MoE & Hybrid linear attention & MoE \\
8 & Qwen3.8 & Hybrid linear attention & Dense \\
9 & GLM-4.7-Flash~\citep{Zeng2025GLM45AR} & MLA & MoE \\
10 & Gemma 4 Dense~\citep{Abd2026Gemma4T} & Sliding-window / global attention & Dense \\
11 & Gemma 4 MoE~\citep{Abd2026Gemma4T} & Sliding-window / global attention & MoE \\
12 & Llama 3~\citep{Dubey2024TheL3} & GQA & Dense \\
13 & Llama 3.1~\citep{Dubey2024TheL3} & GQA & Dense \\
14 & MiniMax-M2.7~\citep{Chen2026TheMS} & GQA & MoE \\
\bottomrule
\end{tabularx}
\end{table}

\begin{table}[t]
\centering
\renewcommand{\arraystretch}{1.4}
\caption{\textbf{Supported sparse methods grouped into four categories.} The inventory includes 15 cache methods and FlashPrefill-v2 for sparse prefill. Each method is assigned to one category according to its primary mechanism.}
\label{tab:supported-sparse-methods}
\small
\begin{tabularx}{0.99\textwidth}{lXl}
\toprule
\textbf{No.} & \textbf{Method} & \textbf{Sparsity Category} \\
\midrule
1 & Quest~\citep{Tang2024QuestQS} & Dynamic Sparsity \\
2 & OmniKV~\citep{Hao2025OmniKVDC} & Dynamic Sparsity \\
3 & RetroInfer~\citep{Chen2025RetroInferAV} & Dynamic Sparsity \\
4 & FlashPrefill-v2~\citep{Fan2026FlashPrefillVB} & Dynamic Sparsity \\
5 & StreamingLLM~\citep{Xiao2023EfficientSL} & KV Eviction \\
6 & SnapKV~\citep{Li2024SnapKVLK} & KV Eviction \\
7 & H$_2$O~\citep{Zhang2023H2OHO} & KV Eviction \\
8 & PyramidKV~\citep{Cai2024PyramidKVDK} & KV Eviction \\
9 & R-KV~\citep{Cai2025RKVRK} & KV Eviction \\
10 & SkipKV~\citep{Tian2025SkipKVSS} & KV Eviction \\
11 & KVzip~\citep{Kim2025KVzipQK} & KV Eviction \\
12 & Palu~\citep{Chang2024PaluCK} & KV Compression \\
13 & DeltaKV~\citep{Hao2026DeltaKVRK} & KV Compression \\
14 & KIVI~\citep{Liu2024KIVIAT} & KV Quantization \\
15 & TurboQuant~\citep{Zandieh2025TurboQuantOV} & KV Quantization \\
16 & FP8 KV & KV Quantization \\
\bottomrule
\end{tabularx}
\end{table}

\subsection{Method Coverage in Figure~\ref{fig:system-comparison}}
\label{app:method-coverage}

\paragraph{Selection Scope.}
Figure~\ref{fig:system-comparison} groups representative implementations by their primary mechanism.
The inventories below provide the broader implementation context for this selection.

\paragraph{Baseline Inventories.}
Vortex revision \texttt{ab9ac68} implements block-sparse routing associated with Kinetics, Quest, and LServe routing~\citep{Chen2026VortexEA}.
The figure selects Quest and LServe routing from these integrations.
Its inherited FP8 KV support is shown under KV Quantization.

SPIN integrates ShadowKV, RetroInfer, and SeerAttention-R, as reported in Section 6.1 of its paper~\citep{Zhao2026UnifyingSA_SPIN}.

Tangram revision \texttt{6fa551f} provides SnapKV, KeyDiff, TOVA, Expected Attention, StreamingLLM, and FastKVzip~\citep{Kim2026TangramUN}.
Its configurable budget scopes are not counted as additional methods.

SparseFrontier revision \texttt{8613c77} provides Vertical-Slash, Block-Sparse, FlexPrefill, and Quest under Dynamic Sparsity, and SnapKV, Ada-SnapKV, and TOVA under KV Eviction~\citep{Nawrot2025TheSF}.
Ada-SnapKV has a separate implementation in its registry.

\paragraph{Serving Capabilities.}
The icons in Figure~\ref{fig:system-comparison} report system-level support.
Vortex inherits batched serving and prefix caching from its SGLang integration.
SPIN supports online batched serving with admission into running batches but does not support prefix caching.
Tangram supports continuous batching but disables prefix caching with compression or ragged paging.
SparseFrontier restricts sparse evaluation to batch size one and disables prefix caching.
SparseEngine supports all listed capabilities, with method-specific restrictions defined in the method registry.

\section{Abstraction Compatibility Criteria}
\label{app:abstraction-compatibility}

\begin{table}[t]
\centering
\small
\setlength{\tabcolsep}{4pt}
\renewcommand{\arraystretch}{1.4}
\caption{\textbf{Abstraction-level compatibility with representative sparse methods.} \yes: the method's defining computation and KV-state semantics fit the documented extension boundary. $\triangle$: only part of the method fits. --: a mapping is not established by that contract. These represent architectural assessments based on documented system boundaries. New method modules are permitted, while system-owned storage and transport contracts remain fixed. Physical eviction must release allocated KV storage instead of merely masking reads. Appendix~\ref{app:abstraction-compatibility} details mapping criteria and evidence.}
\label{tab:abstraction-compatibility}
\resizebox{0.99\textwidth}{!}{%
\begin{tabular}{lcc ccc ccc}
\toprule
\multirow{2.5}{*}{\textbf{System}} & \multicolumn{2}{c}{\textbf{Query-Dep. Retrieval}} & \multicolumn{3}{c}{\textbf{Physical KV Eviction}} & \multicolumn{3}{c}{\textbf{Compressed KV Reps.}} \\
\cmidrule(lr){2-3}\cmidrule(lr){4-6}\cmidrule(l){7-9}
& \textbf{Quest} & \textbf{RetroInfer} & \textbf{SnapKV} & \textbf{Ada-SnapKV} & \textbf{H\textsubscript{2}O} & \textbf{Palu} & \textbf{LoRC} & \textbf{KIVI} \\
\midrule
\textbf{HiSparse} & \yes & $\triangle$ & $\triangle$ & $\triangle$ & $\triangle$ & -- & -- & -- \\
\textbf{Tangram} & -- & -- & \yes & \yes & $\triangle$ & -- & -- & -- \\
\textbf{Vortex} & \yes & $\triangle$ & $\triangle$ & $\triangle$ & $\triangle$ & -- & -- & -- \\
\textbf{SPIN} & \yes & \yes & $\triangle$ & $\triangle$ & $\triangle$ & -- & -- & -- \\
\rowcolor{tableOurs}
\textbf{SparseEngine (Ours)} & \yes & \yes & \yes & \yes & \yes & \yes & \yes & \yes \\
\bottomrule
\end{tabular}}
\end{table}

\paragraph{Scope of the Assessment.}
Table~\ref{tab:abstraction-compatibility} evaluates architectural extensibility through each system's documented extension boundary.
A compatible mapping may add algorithm modules, custom kernels, or method-owned state behind that boundary.
However, it cannot alter storage layouts, allocators, or transport contracts assigned to the shared engine.
A checkmark requires a mapping for the method's defining computation, state updates, and storage semantics.
A triangle identifies a partial mapping, such as selecting the same tokens without physically reclaiming the discarded KV, or providing eviction without the required online score updates.

\paragraph{Method Requirements.}
Quest requires persistent page summaries and query-dependent selection.
RetroInfer additionally requires variable-sized clusters, GPU-CPU retrieval, and attention estimation over unselected content~\citep{Chen2025RetroInferAV}.
SnapKV selects prompt KV using an observation window and physically retains the selected entries. Ada-SnapKV~\citep{Feng2024AdaKVOK} additionally requires non-uniform head budgets.
H$_2$O requires cumulative attention statistics for retained tokens, online updates, and physical eviction as decoding advances.
Palu and LoRC replace full-dimensional KV storage with low-rank representations and matching reconstruction and computation. KIVI requires asymmetric K/V quantization, grouped scales, and a residual full-precision region.
Storing auxiliary summaries alongside an unchanged full KV cache does not implement those representation-level memory savings.

\paragraph{HiSparse.}
The \href{https://www.lmsys.org/blog/2026-04-10-sglang-hisparse}{published design}~\citep{Xie2026HiSparseSS} keeps the complete history in host memory and swaps selected entries into a GPU hot buffer.
For algorithm extensibility we also inspect SGLang revision \texttt{f618022}, including its \href{https://github.com/sgl-project/sglang/blob/f618022b73f813daa6f25e615b0269c82c66a941/python/sglang/srt/mem_cache/sparsity/algorithms/base_algorithm.py}{sparse-algorithm interface}: algorithms construct/update representations and return selected indices, which a backend adaptor maps to attention metadata.
Quest fits this selected-index design. This assessment does not assert that every Quest/HiSparse deployment path is implemented.
Token-retention methods can supply selection policies, but hot-buffer LRU replacement does not delete their host-backed history.
RetroInfer's selection component fits, while its cluster-aware transport and estimation require additional contracts.
The inspected interfaces also do not establish method-owned Palu/LoRC payloads or KIVI's packed payload and quantization-state lifecycle.

\paragraph{Tangram.}
Tangram's scorer, budget-scope, and eviction abstractions support SnapKV and Ada-SnapKV directly~\citep{Kim2026TangramUN}.
At revision \texttt{6fa551f}, the \href{https://github.com/aiha-lab/TANGRAM/blob/6fa551fc8f6edcc118a2a39b3554ee1520e91edd/vllm/v1/attention/compression/qk_scorer_base.py}{scorer contract} is stateless; cached-position rescoring exposes cached keys/values, rather than the current query and accumulated attention state required by H$_2$O.
Its eviction regime supplies retention machinery, so H$_2$O receives a partial mark.
The compression boundary chooses which KV entries remain. It does not expose query-time retrieval of discarded entries or replacement of the retained KV payload with method-defined low-rank or quantized storage.

\paragraph{Vortex.}
Vortex exposes cache-side auxiliary computation and query-side page selection~\citep{Chen2026VortexEA}.
Its \href{https://github.com/Infini-AI-Lab/vortex_torch/blob/ab9ac68c7b9b81f6ba17752741f9e9d92444bf56/vortex_torch/flow/flow.py}{vFlow contract} (revision \texttt{ab9ac68}) reserves the standard K/V entries; \texttt{create\_cache} declares additional tensors, and \texttt{forward\_indexer} writes routing indices for a paged attention backend.
This supports Quest and parts of retention/statistics policies, including the paper's H$_2$O example, without exposing physical reclamation of the standard KV allocation.
RetroInfer additionally needs cluster-aware storage/movement and its estimation operator.
Backend FP8 or MLA support does not by itself establish an extension path for Palu, LoRC, or KIVI's distinct storage semantics.

\paragraph{SPIN.}
Sections 4-5 of SPIN~\citep{Zhao2026UnifyingSA_SPIN} expose algorithm-defined Index, Select, and Attention, with system-managed Offload and Retrieve.
A partition may represent one token, a page, or a variable-sized cluster.
Quest maps to page summaries and selection. RetroInfer is explicitly integrated, including its custom estimation-aware attention.
The same selection interface covers parts of retention policies, but the documented pipeline retrieves subsets of a full tiered KV history and does not establish permanent algorithm-driven deletion.
Its system-owned head-wise KV pages and transport also do not establish a payload-replacement contract for the three representation methods.

\paragraph{SparseEngine.}
The mappings follow the lifecycle boundary and implementation interfaces at revision \texttt{8ea8f80}.
Quest uses page summaries and query-aware compute views.
A RetroInfer integration places cluster maps, summaries, and tiered payloads behind a method-owned cache manager, with a specialized attention provider and explicit memory requirements.
SnapKV, Ada-SnapKV, and H$_2$O place retained indices, head budgets, and persistent scores under the runtime and cache-manager lifecycle, including physical compaction and release.
Palu and LoRC integrations can own latent storage and reconstruction. KIVI maintains packed K/V, scales, and residual storage.
These mappings illustrate feasibility under the lifecycle boundary rather than claiming that all methods are shipped in the default engine binary.

\end{document}